\documentclass{article}

\usepackage{microtype}
\usepackage{graphicx}
\usepackage{subfigure}
\usepackage{booktabs} 

\usepackage[dvipsnames,svgnames,x11names]{xcolor} 

\usepackage[colorlinks=True,
            linkcolor=Maroon,
            anchorcolor=Maroon,  
            pagebackref,
            urlcolor=icmlblue,
            ]{hyperref}

\usepackage[accepted]{icml2023}

\usepackage{amsmath}
\usepackage{amssymb}
\usepackage{mathtools}
\usepackage{amsthm}

\usepackage[capitalize]{cleveref}

\theoremstyle{plain}

\theoremstyle{definition}

\theoremstyle{remark}

\usepackage{microtype}
\usepackage{graphicx}
\usepackage{subfigure}
\usepackage{booktabs} 

 \makeatletter
\def\@fnsymbol#1{\ensuremath{\ifcase#1\or \dagger\or \ddagger\or
  \mathsection\or \mathparagraph\or \|\or **\or \dagger\dagger
  \or \ddagger\ddagger \else\@ctrerr\fi}}
\makeatother

\usepackage{multirow}
\usepackage{amsthm,amsmath,amssymb,lipsum}
\usepackage{mathrsfs}
\usepackage{enumerate}
\usepackage{colortbl}
\usepackage{enumitem}
\usepackage{wrapfig}
\usepackage{bbding}
\usepackage{pifont}
\usepackage{wasysym}
\usepackage{textcomp}

\usepackage{color}
\def\eg{{\it{e.g.}}}

\def\ie{{\it{i.e.}}}

\def\recon{{\scshape ReCon}}

\definecolor{pretty-blue}{RGB}{0, 113, 188}
\definecolor{linecolor}{gray}{.91} 
\definecolor{linecolor2}{gray}{.95} 
\definecolor{linecolor1}{gray}{.97} 

\definecolor{reconcolor}{HTML}{412F8A}
\definecolor{vitcolor}{HTML}{fc8e62}
\newcommand{\reconcolor}[1]{\textcolor{reconcolor}{#1}}
\newcommand{\vitcolor}[1]{\textcolor{vitcolor}{#1}}
\newcommand{\br}{\reconcolor{$\bullet$\,}} 
\newcommand{\bv}{\vitcolor{$\bullet$\,}}  
\newcommand{\bs}{\vitcolor{$\mathbf{\circ}$\,}} 
\newcommand{\bh}{\reconcolor{$\mathbf{\circ}$\,}} 

\icmltitlerunning{~ \hfill \recon~: Contrast with Reconstruct \hfill \thepage}
\begin{document}

\twocolumn[
\icmltitle{
Contrast with Reconstruct: Contrastive 3D Representation Learning \\Guided by Generative Pretraining}

\icmlsetsymbol{equal}{$\dagger$}
\icmlsetsymbol{intern}{$\spadesuit$}

\begin{icmlauthorlist}
\icmlauthor{Zekun Qi}{equal,xjtu}
\icmlauthor{Runpei Dong}{equal,xjtu,intern}
\icmlauthor{Guofan Fan}{xjtu}
\icmlauthor{Zheng Ge}{megvii}
\icmlauthor{Xiangyu Zhang}{megvii}
\icmlauthor{Kaisheng Ma}{thu}
\icmlauthor{Li Yi}{thu,shai,qizhi}
\end{icmlauthorlist}

\icmlaffiliation{xjtu}{Xi'an Jiaotong University}
\icmlaffiliation{megvii}{MEGVII Technology}
\icmlaffiliation{thu}{Tsinghua University}
\icmlaffiliation{shai}{Shanghai AI Laboratory}
\icmlaffiliation{qizhi}{Shanghai Qi Zhi Institute}

\icmlcorrespondingauthor{Kaisheng Ma}{kaisheng@mail.tsinghua.edu.cn}
\icmlcorrespondingauthor{Li Yi}{ericyi@mail.tsinghua.edu.cn}

\icmlkeywords{Machine Learning, ICML, Representation Learning, GPT, Contrastive Learning, Generative Modeling, 3D Point Clouds}

\vskip 0.3in
]



\printAffiliationsAndNotice{\icmlEqualContribution \icmlInternNotice} 

\begin{abstract}
    Mainstream 3D representation learning approaches are built upon contrastive or generative modeling pretext tasks, where great improvements in performance on various downstream tasks have been achieved. However, we find these two paradigms have different characteristics:
    (i) contrastive models are data-hungry that suffer from a \textit{representation over-fitting} issue;
    (ii) generative models have a \textit{data filling} issue that shows inferior data scaling capacity compared to contrastive models.
    This motivates us to learn 3D representations by sharing the merits of both paradigms, which is non-trivial due to the \textit{pattern difference} between the two paradigms.
    In this paper, we propose \textit{Contrast with Reconstruct} (\recon) that unifies these two paradigms.
    \recon~is trained to learn from both generative modeling teachers and single/cross-modal contrastive teachers through ensemble distillation, where the generative student guides the contrastive student.
    An encoder-decoder style \recon-block is proposed that 
    transfers knowledge through cross attention with stop-gradient, which avoids pretraining over-fitting and pattern difference issues.
    \recon~achieves a new state-of-the-art in 3D representation learning, \eg, \textbf{91.26}\% accuracy on ScanObjectNN.
    Codes have been released at \url{https://github.com/qizekun/ReCon}.
\end{abstract}

\section{Introduction}\label{sec:intro}
\vspace{-2pt}
Self-supervised representation learning (SSRL) has witnessed a booming era of \textit{foundational models}~\citep{FoundationModel21}, significant advancements are being made in natural language processing (NLP)~\citep{GPTv1_18,BERT,GPT3_20,CoT22,InstructGPT22}, 2D machine vision~\citep{MoCo,MAE}, and both (vision-language, VL)~\citep{CLIP,StableDiffusion22,Flamingo22}.
While this great course toward foundational machine intelligence is trending, the success of these methods generally demands training on data of \textit{extreme} size.
However, compared to 2D vision and NLP, 3D vision is faced with a challenging \textit{data desert} issue~\citep{ACT23}
due to collection difficulty.

\begin{figure}[t!]
    \begin{center}
    \vspace{-2pt}
    \includegraphics[width=\linewidth]{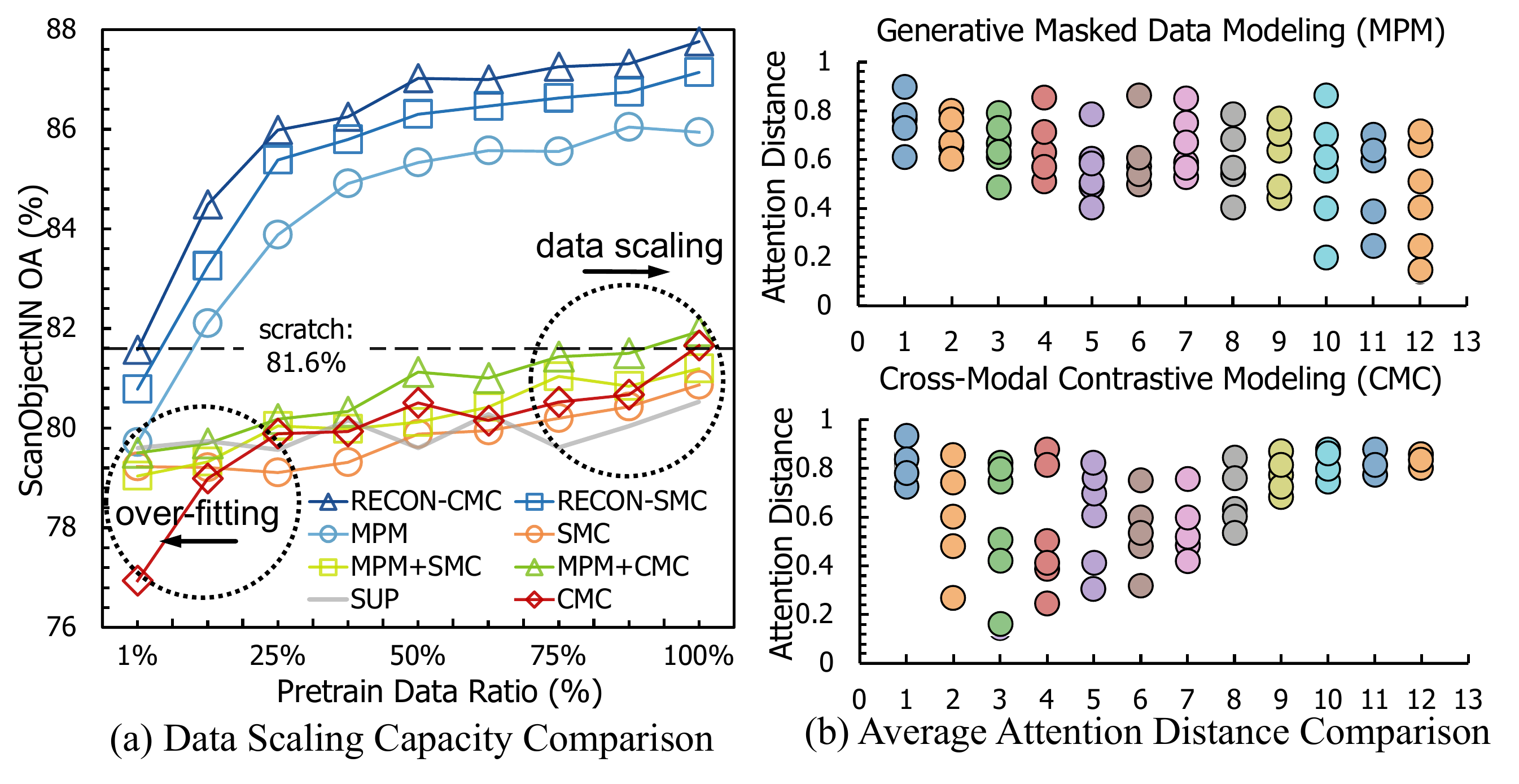}
    \vspace{-22pt}
    \caption{\textbf{Data efficiency comparison and attention distance visualization}. (a) Fine-tuning overall accuracy on ScanObjectNN PB\_T50\_RS with models pretrained with different methods on ShapeNet of different data ratios. (b) Averaged attention distance of models pretrained on ShapeNet with \textit{generative} masked point modeling (MPM)~\citep{PointMAE} and cross-modal \textit{contrastive} (CMC) modeling~\citep{CLIP} (texts, images, and point clouds are used), respectively. SUP: supervised classification pretraining on ShapeNet. SMC: single-modal contrastive pretraining~\citep{SCL20}. \recon-SMC and \recon-CMC represent our proposed \recon~with single-modal and cross-modal contrastive modeling variants, respectively. MPM+SMC and MPM+CMC represent vanilla multi-task learning.
    }\label{fig:data_scaling}
    \vspace{-24pt}
    \end{center}
\end{figure}
\begin{figure*}[t!]
    \begin{center}
    \vspace{-5pt}
    \includegraphics[width=0.90\linewidth]{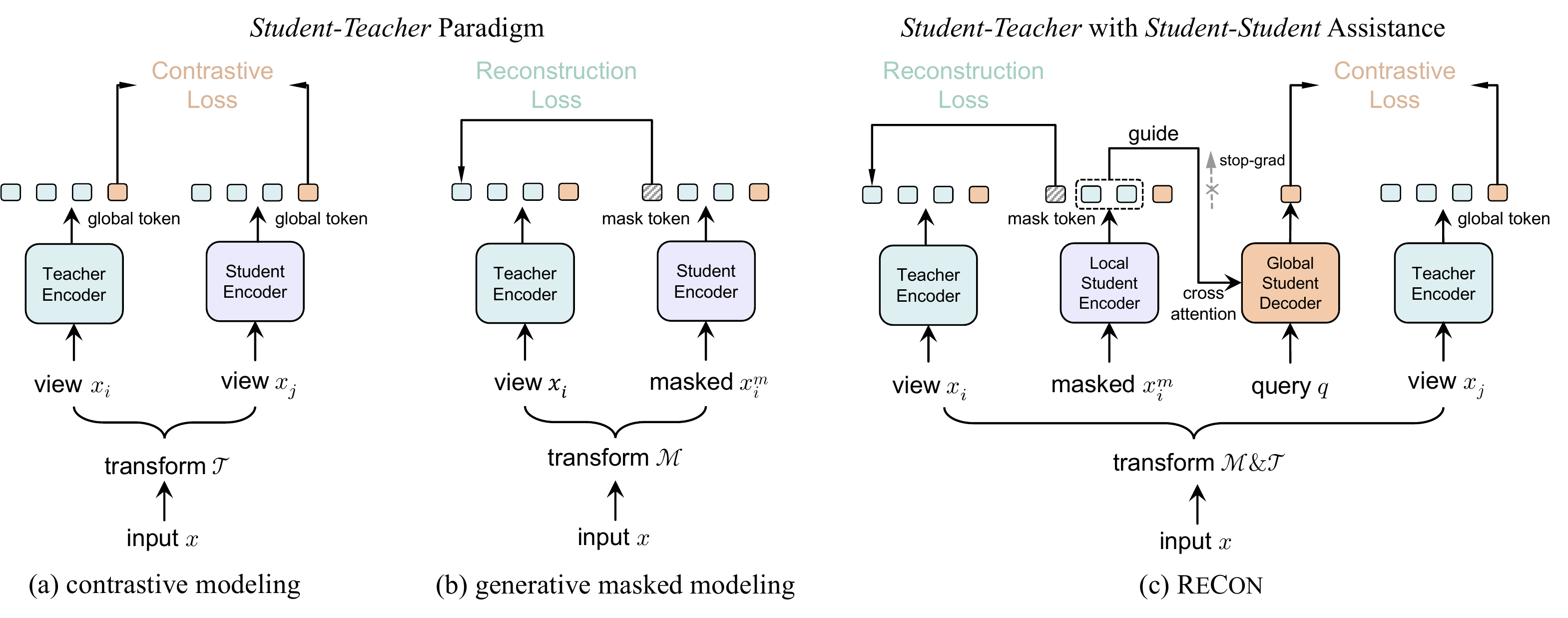}
    \vspace{-12pt}
    \caption{\textbf{Concept comparison of \textit{contrastive}, \textit{generative}, and our \recon~paradigms}. Here, we illustrate the methods in a unified view of knowledge distillation (see Sec.~\ref{sec:kd_view}). (a) Contrastive students are trained to learn \textit{invariance} from the teacher. (b) Generative masked modeling encourages students to reconstruct clean signals provided by the teacher. (c) \recon~unifies the two paradigms by learning from multi-teachers, where the generative local student is also a ``teacher" that guides the contrastive global student.
    }\label{fig:concept} 
    \end{center}
    \vspace{-18pt}
\end{figure*}
Though under this \textit{low-data} regime, numerous 3D SSRL methods have been developed, which can be grouped into 
two categories, \ie, \textit{contrastive (single/cross-modal)}~\citep{PointContrast20,DepthContrast21,4DContrast22,TupleInfoNCE21,CrossPoint22} and \textit{generative (reconstruct/predict)}~\citep{OcCo,PointBERT,PointMAE} methods.
However, we investigate the pretraining efficiency of the two paradigms by scaling the pretraining data of ShapeNet ranging from 1\% to 100\%, and we find that these two paradigms have their issues (see \cref{fig:data_scaling}(a)):
\vspace{-10pt}
\begin{itemize}
    \item \textbf{Representation over-fitting (contrastive)}. Contrastive models fail to bring generalization when the pretraining data is lacking ($<90\%$), while generative models bring significant improvements with only $\sim 25\%$ data. It shows that contrastive models can easily find shortcuts with trivial representations that \textit{over-fit} the limited data~\citep{MoCo}, and generative models are less data-hungry that learn decent initialization with very few data~\citep{InvariantMIM22}.
    \vspace{-12pt}
    \item \textbf{Data filling (generative)}. Contrastive models present a better potential with scaled-up data, while generative models only provide a little improvement. 
    It shows that contrastive learning may bring superior \textit{data-scaling} capacity when the pretraining data is sufficient. 
    This is observed in 2D where contrastive models surpass generative models~\citep{TuneCLIP22} that have less scaling capability~\citep{ScaleMIM22}.
\end{itemize}
\vspace{-10pt}

These observations motivate the design that shares both merits without mentioned issues.
However, as shown in \cref{fig:data_scaling}(a), simply combining these two paradigms as multi-task learning leads to unsatisfactory results -- lower than the generative model baseline and the \textit{representation over-fitting} issue remains. 
To understand why, we visualize the averaged attention distance\footnote{The attention distance are conducted by averaging per attention head for each layer following~\citet{ViT}, which reveals the relative receptive field of the learned attention.} of the contrastive and generative models in \cref{fig:data_scaling}(b).
A \textit{pattern difference} issue is observed that the attention of contrastive models is mainly paid to a \textit{global} field, while generative models have an appetite for focused \textit{local} attention, which is consistent with the observations by \citet{DarkMIM22} in 2D. 
We conject that this \textit{pattern difference} issue causes a task conflict in the naive multi-task representation learning setting, indicating it is \textit{non-trivial} to combine the merits of contrastive and generative modeling.

To address the above-mentioned issues, we propose \textit{Contrast with Reconstruct} (\recon) that trains generative modeling as guidance for contrastive learning while sharing both merits.
As shown in \cref{fig:concept}, from the perspective of knowledge distillation (Sec.~\ref{sec:kd_view}), contrastive and generative methods can be viewed as vanilla \textit{student-teacher} paradigms (\cref{fig:concept}(a-b)). 
In contrast, our \recon\ unifies the two paradigms as ensemble distillation from multi-teachers, while the generative student is also a ``teacher" which guides the contrastive learning (\textit{student-teacher} knowledge distillation with \textit{student-student} assistance).

In particular, inspired by~\citet{AttentionIsAllYouNeed}, a novel encoder-decoder style \recon-block is proposed where cross attention with stop-gradient is used to transfer guidance from reconstruction to contrastive modeling.
In this fashion, the knowledge from multi-teachers is disentangled and reinforced, which addresses the \textit{representation over-fitting} issue and avoids the learned \textit{pattern difference}.
Further, our \recon~utilizes \textit{single-modal} or \textit{cross-modal} contrastive learning of 3D point clouds, 2D RGB images, and languages that significantly enlarge the pretraining data diversity and capacity. 
Meanwhile, the \textit{data-filling} issue of the generative student is alleviated due to the promising scaling capacity of the contrastive student.
\cref{fig:data_scaling}(b) shows that our \recon~learns 3D representations with high generalization capacity.
By transferring the learned representations to various benchmarks, a new state-of-the-art in self-supervised 3D learning is achieved. For example, an average improvement of \textbf{+9.2\%} and \textbf{+2.9\%} accuracy are achieved on ScanObjectNN and ModelNet40, respectively.
These results show that \recon\ learns foundational geometric understanding, and this simple and general framework successfully unifies contrastive and generative modeling.
\section{Related Works}
\textbf{Contrastive Representation Learning}~
is one of the mainstream self-supervised learning paradigms~\citep{LeCunContrastive}, which learns potential semantics from constructed \textit{invariance} or \textit{equivaiance}~\citep{EquivariantSSL22}.
Generally, Instance Discrimination~\citep{InstanceDiscrimination18} is widely adopted to align and distinguish representations of views with the same high-level semantics or not. 
The views could be constructed by augmentations to single-modal~\citep{SimCLR,MoCo,MoCoThree21} or multi-modal data~\citep{CLIP,GLIP22}.
Most works use global features for processing. 
For example, SimCLR~\citep{SimCLR} uses samples with different augmentation policies to construct positive and negative pairs. 
CLIP~\citep{CLIP} proposes a two-tower network that aligns the global representation of languages and images. 
Driven by \textit{InforMAX principle}~\citep{DeepInfoMax19}, they generally use InfoNCE~\citep{InfoNCE} as the loss function to maximize mutual information. 
In 3D, PointContrast~\citep{PointContrast20} proposes geometric augmentation to generate positive and negative pairs. 
CrossPoint~\citep{CrossPoint22} uses both inter and intra-modal contrastive learning. 
PointCLIP~\citep{PointCLIP22} realizes image-point alignment by projecting point clouds to 2D depth images. 
In this work, we focus on single/cross-modal contrastive learning by discriminative contrast~\citep{SCL20} or global feature alignment like~\citet{CLIP}, which is guided by masked generative modeling.

\textbf{Generative Masked Representation Learning}~ 
has emerged as another paradigm of self-supervised learning from NLP~\citep{BERT} to Vision~\citep{MAE}.
It requires the models to learn structured knowledge by \textit{reconstructing masked} input data, which encourages the association of different local patches.
In NLP, it has been a dominant approach to probing knowledge by recovering or predicting words in sentences~\citep{BERT,GPT3_20}. 
With the rapid development of Transformers in vision~\citep{ViT,SwinT}, abundant works have been proposed to realize mask image modeling (MIM). 
\citet{MAE} propose masked autoencoder (MAE) to reconstruct RGB pixels. 
\citet{BEiT} reconstructs the VQVAE~\citep{DALL-E} codebook with encoded semantics. 
Some works propose to reconstruct online teacher tokens~\citep{iBoT} or HOG features~\citep{MaskFeat}. 
In 3D, PointMAE~\citep{PointMAE} extends MAE~\citep{MAE} by reconstructing masked point clouds. PointM2AE~\citep{PointM2AE22} uses a hierarchical Transformer and designs the corresponding masking strategy. MaskPoint~\citep{MaskPoint} proposes to add some noise points and classify whether they belong to masking tokens. Recently, ACT~\citep{ACT23} uses a cross-modal autoencoder as the reconstruction target to acquire dark knowledge from other modalities. 
\section{\recon : Contrast with Reconstruct}\label{sec:method}
\begin{figure}[t!]
    \begin{center}
    \includegraphics[width=0.9\linewidth]{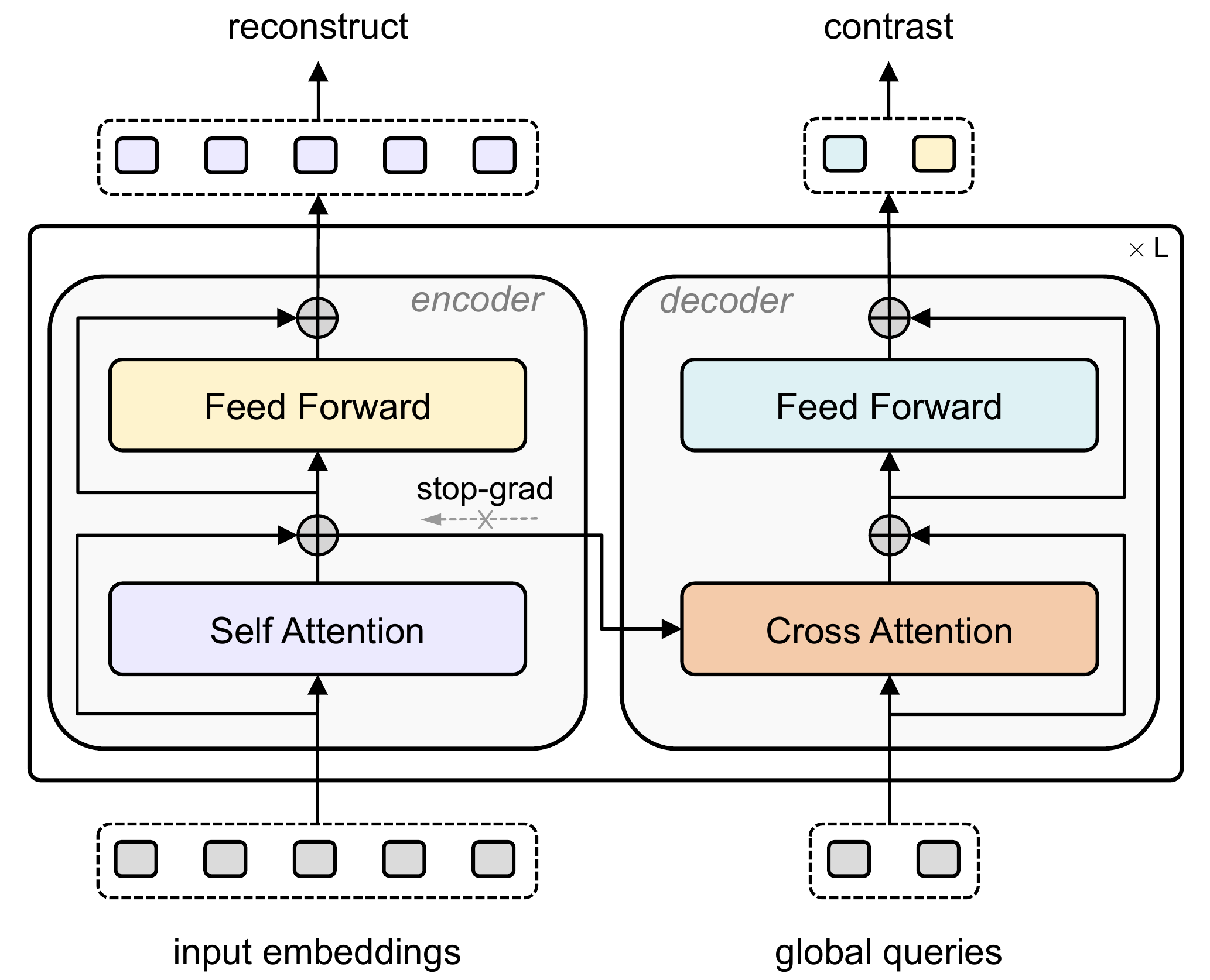}
    \vspace{-5pt}
    \caption{\textbf{Illustration of the proposed \recon-block for pretraining}. The \textit{local} reconstruction task is used to train the encoder, while the \textit{global} contrastive task is used to train the decoder guided by reconstruction-oriented embeddings with cross attention (CA). Stop-gradient (stop-grad) is applied to every CA connection to avoid misleading gradient flow from \textit{global} to \textit{local}.}\label{fig:recon_block}
    \end{center}
    \vspace{-10pt}
\end{figure}
We begin with a review in a unified view of knowledge distillation for the two mainstream representation learning methods: masked \textit{generative} modeling and \textit{contrastive} modeling. We then introduce \recon~that unifies these two representation learning methods by reconstruction-guided contrastive learning using an encoder-decoder style \recon-block based architecture, where the overall representation learning is formulated as distillation with both teachers with student-student assistance.

\begin{figure*}[ht!]
    \begin{center}
    \vspace{-4pt}
    \includegraphics[width=0.84\linewidth]{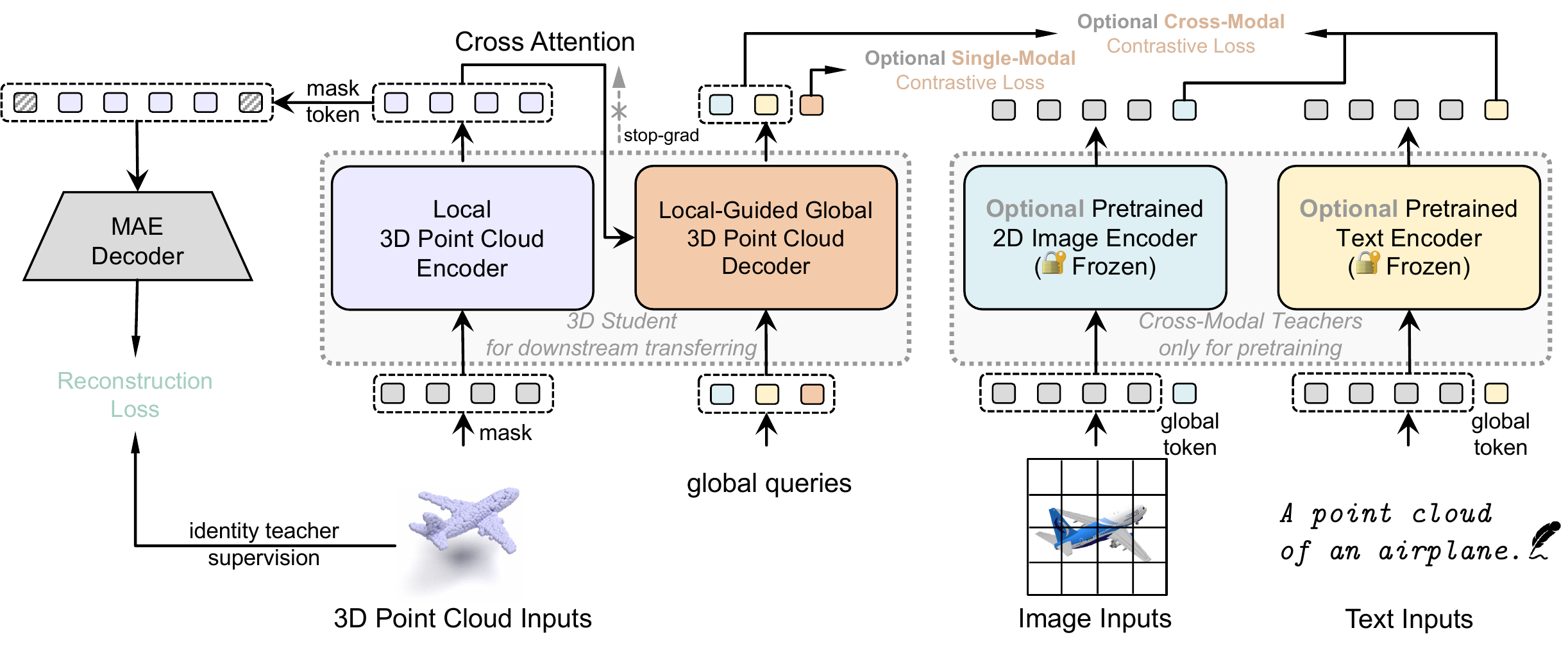}
    \vspace{-12pt}
    \caption{\textbf{Overview of \recon}. \recon\ can be applied to \textit{either} single-modal 3D point clouds inputs \textit{or} cross-modal inputs with rendered RGB images and text descriptions, which will be encoded as sequential tokens. The 3D token embeddings are then masked for \textit{generative} reconstruction for the local 3D encoder, where the encoded intermediate embeddings are fed to the global 3D decoder with stop-gradient (stop-grad) through cross-attention. The \textit{global queries} are learnable and supervised by global \textit{contrastive} learning. 
    }\label{fig:framework}
    \vspace{-15pt}
    \end{center}
\end{figure*}
\subsection{Knowledge Distillation: A Unified View of Generative and Contrastive Learning}\label{sec:kd_view}
\paragraph{Contrastive Modeling}
The key insight of contrastive learning lies in \textit{invariance learning}~\citep{InvarianceTheory11,MoCo,InvariantMIM22}, where the abstraction of semantics is generally invariant or equivariant~\citep{EquivariantSSL22} to multiple transformed views like augmentations~\citep{SimCLR} or modalities~\citep{CMC20}.
From the perspective of knowledge distillation~\citep{HintonKD15}, it can be viewed as a student network learning the \textit{invariance} knowledge transferred from the encoded views of the teacher. 
Formally, given input data $x\sim \mathcal{D}$ with distribution $\mathcal{D}$, the student network is $\mathcal{F}^{{S}}_\theta(\cdot)$ with parameters $\theta$ and $\mathcal{F}^{{T}}_\phi(\cdot)$ is the teacher network with parameters $\phi$. The optimization target can be written:
\begin{equation}\label{eq:kd_contrast}
\begin{aligned}
    \mathop{\min}\limits_{\theta} \mathop{\mathbb{E}}\limits_{\mathop{x \sim \mathcal{D},}\limits_{\{\mathcal{T}_i,\mathcal{T}_j\}\in \mathcal{T}}} 
    & {\mathcal{L}}^{\text{CON}}_{\mathbb{C}} (z_i, z_j), \\
    & z_i = \mathcal{F}^{{S}}_\theta\big(\mathcal{T}_i(x)\big),
    z_j = \mathcal{F}^{{T}}_\phi\big(\mathcal{T}_j(x)\big),
\end{aligned}
\end{equation}
where
\begin{itemize}
    \item The teacher parameters $\phi$ can be the same as $\theta$ with~\citep{SimSiam} or without~\citep{SimCLR} \textit{stop-gradient}, $\phi$ can also be the \textit{momentum counterpart} updated as $\phi\leftarrow m\phi + (1-m)\theta$ where $m\in(0,1]$ is the momentum coefficient~\citep{MoCo,MoCoThree21}. For multi-modal inputs\footnote{Note that there are some methods~\citep{CMC20,CLIP} that optimizes both $\theta$ and $\phi$.}, $\phi$ can be \textit{different} from $\theta$ that encodes views from other modalities~\citep{CMC20,CLIP,CenterLoss21}.
    \item $\{\mathcal{T}_i(\cdot),\mathcal{T}_j(\cdot)\}\in \mathcal{T}$ are two transformations of the input data that belongs to constructed transformation pairs $\mathcal{T}$, where $\mathcal{T}_j(\cdot)$ constructs the positive \textit{or} negative\footnote{For methods use positive samples only~\citep{BYOL,SimSiam}, $\mathcal{T}_j(\cdot)$ only generates positive views.} invariance targets, respectively.
    In general, this transformation generates copied views by data augmentations~\citep{DINO21,StrongAugContrast21}, we extend it to \textit{multi-modal} inputs where the transformation becomes generating views from different modalities that share the same high-order concept~\citep{CLIP,TupleInfoNCE21}.
    \item ${\mathcal{L}}^{\text{CON}}_{\mathbb{C}}(\cdot,\cdot)$ is the distance function defined in some metric space $\mathbb{C}$.
    To avoid representation collapsing~\citep{MoCo}, \textit{InfoMax-Principle}~\citep{DeepInfoMax19,AugDeepInfoMax19} based metrics like MINE~\citep{MINE18}, InfoNCE~\citep{InfoNCE} are often used~\citep{MoCo,GoodViewContrast20,CDS22}. For methods using \textit{positive-only} transformations, the metric function can be feature correlation measurement~\citep{BarlowTwins21}, $\ell_2$ distance~\citep{WhiteContrast21}, or cosine similarity~\citep{BYOL,SimSiam}.
\end{itemize}
\paragraph{Generative Masked Modeling} Similarly, given input data $x\sim \mathcal{D}$ with distribution $\mathcal{D}$, the student network as $\mathcal{F}^{{S}}_\theta(\cdot)$ with parameters $\theta$ and $\mathcal{F}^{{T}}_\phi(\cdot)$ as the teacher network\footnote{Note that for methods like MAE~\citep{MAE,PointMAE}, the teacher network is an identity mapping with no parameters, and therefore $\phi$ is an empty set, see~\citet{ACT23}.} with parameters $\phi$. 
generative masked modeling can be formulated as follows:
\begin{equation}\label{eq:kd_mdm}
\begin{aligned}
    \mathop{\min}\limits_{\theta} \mathop{\mathbb{E}}\limits_{\mathop{x \sim \mathcal{D},}\limits_{\{\mathcal{M}_i,\widetilde{\mathcal{M}}_i\}\in \mathcal{M}}} 
    & {\mathcal{L}}^{\text{REC}}_{\mathbb{D}} (z_i, z_j), \\
    & z_i = \mathcal{F}^{{S}}_\theta\big(\mathcal{M}_i(x)\big), 
    z_j = \mathcal{F}^{{T}}_\phi\big(\widetilde{\mathcal{M}}_i(x)\big),
\end{aligned}
\end{equation}
where 
\begin{itemize}
    \item ${\mathcal{L}}^{\text{REC}}_{\mathbb{D}}$ is a distance function defined in some metric space $\mathbb{D}$. It can be $\ell_2$ distance~\citep{MAE}, cross-entropy~\citep{BERT,BEiT}, or Chamfer-Distance~\citep{ChamferDistance17,PointMAE}.
    \item $\{\mathcal{M}_i,\widetilde{\mathcal{M}}_i\}\in \mathcal{M}$ are the masking corruptions where $\mathcal{M}_i(\cdot)$ samples a subset of the input tokens and performs masking~\citep{BERT}, and $\widetilde{\mathcal{M}}_i(\cdot)$ correspondingly samples the subset while without masking.
\end{itemize}

\begin{table*}[ht!]
\caption{
\textbf{Classification results on the ScanObjectNN and ModelNet40 datasets}. The inference model parameters \#P (M), FLOPS \#F (G), and overall accuracy (\%) are reported. 
The dagger ($^{\dagger}$) denotes that the model was reproduced using our proposed \br\recon-block and the fine-tuning techniques used in \recon.
We compare with methods using the \bh hierarchical Transformer architectures (\eg, Point-M2AE~\citep{PointM2AE22}), \bv plain Transformer architectures, and \bs dedicated architectures for 3D understanding.
\texttt{PT}: pretrained teacher is used, \texttt{MD}: multi-modal data is used. Single-Modal means that \textbf{only point clouds} are used as pre-training data.
}
\label{tab:scanobjectnn}
\begin{center}
\vspace{-14pt}
\resizebox{1.0\linewidth}{!}{
\begin{tabular}{lccccccccc}
\toprule[0.95pt]
\multirow{2}{*}[-0.5ex]{Method} & \multirow{2}{*}[-0.5ex]{\#P} & \multirow{2}{*}[-0.5ex]{\#F} & \multirow{2}{*}[-0.5ex]{\texttt{PT}} & \multirow{2}{*}[-0.5ex]{\texttt{MD}} & \multicolumn{3}{c}{ScanObjectNN} & \multicolumn{2}{c}{ModelNet40}\\
\cmidrule(lr){6-8}\cmidrule(lr){9-10} & & & & & OBJ\_BG & OBJ\_ONLY & PB\_T50\_RS& 1k P & 8k P\\
\midrule[0.6pt]
\multicolumn{10}{c}{\textit{Supervised Learning Only}}\\
\midrule[0.6pt]
\bs PointNet~\citep{PointNet} & 3.5 & 0.5 & $\times$ & $\times$ & 73.3 & 79.2 & 68.0 & 89.2 & 90.8\\
\bs PointNet++~\citep{PointNet++} & 1.5 & 1.7 & $\times$ & $\times$ & 82.3 & 84.3 & 77.9 & 90.7 & 91.9\\
\bs DGCNN~\citep{DGCNN} & 1.8 & 2.4 & $\times$ & $\times$ & 82.8 & 86.2 & 78.1 & 92.9 & -\\
\bs PointCNN~\citep{PointCNN} & 0.6 & - & $\times$ & $\times$ & 86.1 & 85.5 & 78.5 & 92.2 & -\\
\bs SimpleView~\citep{SimpleView} & - & - & $\times$ & $\times$ & - & - & 80.5$\pm$0.3 & 93.9 & -\\
\bs MVTN~\citep{MVTN} & 11.2 & 43.7 & $\times$ & $\times$ & 92.6 & 92.3 & 82.8 & 93.8 & -\\
\bh PCT~\citep{PCT} & 2.88 & 2.3 & $\times$ & $\times$ & - & - & - & 93.2 & -\\
\bs PointMLP~\citep{PointMLP} & 12.6 & 31.4 & $\times$ & $\times$ & - & - & 85.4$\pm$0.3 & 94.5 & -\\
\bs PointNeXt~\citep{PointNext} & 1.4 & 3.6 & $\times$ & $\times$ & - & - & 87.7$\pm$0.4 & 94.0 & -\\
\bs P2P-HorNet~\citep{P2P} & - & 34.6 & $\checkmark$ & $\checkmark$ & - & - & 89.3 & 94.0 & -\\
\midrule[0.6pt]
\multicolumn{10}{c}{\textit{with Single-Modal Self-Supervised Representation Learning} ({\scshape Full})}\\
\midrule[0.6pt]
\bv Transformer~\citep{AttentionIsAllYouNeed} & 22.1 & 4.8 & $\times$ & $\times$ & 83.04 & 84.06 & 79.11 & 91.4 & 91.8\\
\br Transformer$^\dagger$~\citep{AttentionIsAllYouNeed} & 43.6 & 5.3 & $\times$ & $\times$ & 84.90 & 86.12 & 81.64 & 91.6 & 92.0\\
\bv Point-BERT~\citep{PointBERT} & 22.1 & 4.8 & $\times$ & $\times$ & 87.43 & 88.12 & 83.07 & 93.2 & 93.8\\
\bv Point-MAE~\citep{PointMAE} & 22.1 & 4.8 & $\times$ & $\times$ & 90.02 & 88.29 & 85.18 & 93.8 & 94.0\\
\bh Point-M2AE~\citep{PointM2AE22} & 15.3 & 3.6 & $\times$ & $\times$ & 91.22 & 88.81 & 86.43 & 94.0 & -\\
\br Point-MAE$^\dagger$~\citep{PointMAE} & 43.6 & 5.3 & $\times$ & $\times$ & 92.60 & 91.91 & 88.42 & 93.8 & 94.0\\
\rowcolor{linecolor2}\br\recon~\textbf{w/o vot.} & 43.6 & 5.3 & $\times$ & $\times$ & \textbf{94.15} & \textbf{93.12} & \textbf{89.73} & 93.6 & 93.8\\
\rowcolor{linecolor}\br\recon~\textbf{w/ vot.} & 43.6 & 5.3 & $\times$ & $\times$ & \textbf{94.49} & \textbf{93.29} & \textbf{90.35} & \textbf{93.9} & \textbf{94.2}\\
\midrule[0.6pt]
\multicolumn{10}{c}{\textit{with Cross-Modal Self-Supervised Representation Learning} ({\scshape Full})}\\
\midrule[0.6pt]
\bv ACT~\citep{ACT23} & 22.1 & 4.8 & $\checkmark$ & $\times$ & 93.29 & 91.91 & 88.21 & 93.7 & 94.0\\
\rowcolor{linecolor1}\br\recon-Tiny~\textbf{w/o vot.} & 11.4 & 2.4 & $\checkmark$ & $\checkmark$ & \textbf{93.80} & \textbf{92.94} & \textbf{89.10} & 93.3 & 93.6\\
\rowcolor{linecolor2}\br\recon-Small~\textbf{w/o vot.} & 19.0 & 3.2 & $\checkmark$ & $\checkmark$ & \textbf{94.15} & \textbf{93.12} & \textbf{89.52} & 93.5 & 93.8\\
\rowcolor{linecolor1}\br\recon~\textbf{w/o vot.} & 43.6 & 5.3 & $\times$ & $\checkmark$ & \textbf{94.66} & \textbf{93.29} & \textbf{90.32} & \textbf{94.0} & \textbf{94.2}\\
\rowcolor{linecolor2}\br\recon~\textbf{w/o vot.} & 43.6 & 5.3 & $\checkmark$ & $\checkmark$ & \textbf{95.18} & \textbf{93.63} & \textbf{90.63} & \textbf{94.1} & \textbf{94.3}\\
\rowcolor{linecolor}\br\recon~\textbf{w/ vot.} & 43.6 & 5.3 & $\checkmark$ & $\checkmark$ & \textbf{95.35} & \textbf{93.80} & \textbf{91.26} & \textbf{94.5} & \textbf{94.7}\\
\midrule[0.6pt]
\multicolumn{10}{c}{\textit{with Self-Supervised Representation Learning} ({\scshape Mlp-Linear})}\\
\midrule[0.6pt]
\br Point-MAE$^\dagger$~\citep{PointMAE} & 43.6 & 5.3 & $\times$ & $\times$ & 82.77$\pm$0.30 & 83.23$\pm$0.16 & 74.13$\pm$0.21 & 90.22$\pm$0.09 & 90.73$\pm$0.09 \\
\bv ACT~\citep{ACT23} & 22.1 & 4.8 & $\checkmark$ & $\times$ & 85.20$\pm$0.83 & 85.84$\pm$0.15 & 76.31$\pm$0.26 & 91.36$\pm$0.17 & 91.75$\pm$0.18 \\
\rowcolor{linecolor}\br\recon~\textbf{w/o vot.} & 43.6 & 5.3 & $\checkmark$ & $\checkmark$ & \textbf{89.50}$\pm$0.20 & \textbf{89.72}$\pm$0.17 & \textbf{81.36}$\pm$0.14 & \textbf{92.47}$\pm$0.22 & \textbf{92.68}$\pm$0.07\\
\midrule[0.6pt]
\multicolumn{10}{c}{\textit{with Self-Supervised Representation Learning} ({\scshape Mlp-3})}\\
\midrule[0.6pt]
\br Point-MAE$^\dagger$~\citep{PointMAE} & 43.6 & 5.3 & $\times$ & $\times$ & 85.78$\pm$0.31 & 85.51$\pm$0.16 & 80.38$\pm$0.21 & 91.25$\pm$0.24 & 91.68$\pm$0.19 \\
\bv ACT~\citep{ACT23} & 22.1 & 4.8 & $\checkmark$ & $\times$ & 87.14$\pm$0.22 & 87.90$\pm$0.40 & 81.52$\pm$0.19 & 92.69$\pm$0.18 & 92.95$\pm$0.10 \\
\rowcolor{linecolor}\br\recon~\textbf{w/o vot.} & 43.6 & 5.3 & $\checkmark$ & $\checkmark$ & \textbf{90.62}$\pm$0.22 & \textbf{90.71}$\pm$0.30 & \textbf{83.80}$\pm$0.42& \textbf{93.00}$\pm$0.10 & \textbf{93.39}$\pm$0.05\\
\bottomrule[0.95pt]
\end{tabular}
}
\end{center}
\vspace{-16pt}
\end{table*}

\textbf{Ensemble Representation Distillation}~~
Ensemble distillation\footnote{Term taken from~\citet{CRD20}.} is known to be more informative and instructive~\citep{CRD20,Feed20,MultiExpertKD,KDOverview22}, which encourages the student network to learn ensembled and disentangled knowledge representation. We motivate our method as an ensembled knowledge distillation of Eq.~(\ref{eq:kd_contrast}) and Eq.~(\ref{eq:kd_mdm}), where the student is trained with merits from both \textit{contrastive} and \textit{generative} aspects. The overall loss $\mathcal{L}^{\text{\recon}}$ is defined as follows:
\begin{equation}\label{eq:ensemble_kd}
    \mathcal{L}^{\text{\recon}} = \mathcal{L}^{\text{REC}} + \mathcal{L}^{\text{CON}}.
\end{equation}

\subsection{Reconstruction Guided Contrastive Learning}\label{sec:method_impl}
\vspace{-6pt}
\textbf{Network Architecture}~~ As discussed in Sec.~\ref{sec:intro}, since the two distillation result in different learning patterns, it is \textit{non-trivial} to learn from both targets jointly. 
To tackle this issue, we propose \textit{contrast with reconstruct} (\recon). The reconstruction-oriented representations that focus on \textit{local} patterns are used as semantic guidance for \textit{global} contrastive learning. 
Inspired by the Transformer encoder-decoder architecture~\citep{AttentionIsAllYouNeed}, we conduct dense masked modeling with an encoder, which produces features to guide global contrastive learning through a sparse query-based decoder. 
The encoder and decoder share the same Transformer architecture, and they are layer-wisely associated with cross attention (CA). 
Due to limited 3D data, the contrastive model can easily learn trivial representations as shortcuts, and this could lead to noisy training signals, which may harm generative student learning.
Hence, to avoid the task conflicts between these two students, we use \textit{stop-gradient} for every CA connection to cut the misleading training signal from global contrast to local reconstruction.
We call the proposed network architecture \recon-block, which is illustrated in~\cref{fig:recon_block}. 
In this fashion, the ensemble ``student-teacher" distillation from multi-teacher is learned jointly with ``student-student" assistance where the contrastive student is guided by the generative ``teacher".
As a result, the contrastive student is trained with a good data scaling capacity without the risk of representation over-fitting, and the pattern difference issue is avoided with no task conflicts.
\begin{center}
\begin{table}[t!]
    \vspace{-2.5pt}
    \centering
    \setlength\tabcolsep{8pt}
    \caption{\textbf{Few-shot classification results on ModelNet40}. $^\dagger$ represent results of our proposed \br\recon-block built backbone architecture. Overall accuracy (\%) without voting is reported.}\label{tab:few-shot}
    \resizebox{\linewidth}{!}{
    \begin{tabular}{lcccc}
    \toprule[0.95pt]
    \multirow{2}{*}[-0.5ex]{Method}& \multicolumn{2}{c}{5-way} & \multicolumn{2}{c}{10-way}\\
    \cmidrule(lr){2-3}\cmidrule(lr){4-5} & 10-shot & 20-shot & 10-shot & 20-shot\\
    \midrule[0.6pt]
    \bs DGCNN &31.6 $\pm$ 2.8 &  40.8 $\pm$ 4.6&  19.9 $\pm$  2.1& 16.9 $\pm$ 1.5\\
    \bs OcCo &90.6 $\pm$ 2.8 & 92.5 $\pm$ 1.9 &82.9 $\pm$ 1.3 &86.5 $\pm$ 2.2\\
    \midrule[0.6pt]
    \multicolumn{5}{c}{\textit{with Self-Supervised Representation Learning} ({\scshape Full})}\\
    \midrule[0.6pt]
    \bv Transformer & 87.8 $\pm$ 5.2& 93.3 $\pm$ 4.3 & 84.6 $\pm$ 5.5 & 89.4 $\pm$ 6.3\\
    \br Transformer$^\dagger$ & 90.2 $\pm$ 5.9 & 94.3 $\pm$ 4.4 & 85.2 $\pm$ 5.9 & 89.9 $\pm$ 6.1\\
    \bs OcCo & 94.0 $\pm$ 3.6& 95.9 $\pm$ 2.3 & 89.4 $\pm$ 5.1 & 92.4 $\pm$ 4.6\\
    \bv Point-BERT & 94.6 $\pm$ 3.1 & 96.3 $\pm$ 2.7 &  91.0 $\pm$ 5.4 & 92.7 $\pm$ 5.1\\
    \bv MaskPoint & 95.0 $\pm$ 3.7 & 97.2 $\pm$ 1.7 & 91.4 $\pm$ 4.0 & 93.4 $\pm$ 3.5\\
    \bv Point-MAE & 96.3 $\pm$ 2.5&97.8 $\pm$ 1.8 & 92.6 $\pm$ 4.1 & 95.0 $\pm$ 3.0\\
    \bh Point-M2AE & 96.8 $\pm$ 1.8&98.3 $\pm$ 1.4 & 92.3 $\pm$ 4.5 & 95.0 $\pm$ 3.0\\
    \br Point-MAE$^\dagger$ & 96.4 $\pm$ 2.8&97.8 $\pm$ 2.0 & 92.5 $\pm$ 4.4 & 95.2 $\pm$ 3.9\\
    \bv ACT & 96.8 $\pm$ 2.3 & 98.0 $\pm$ 1.4 & 93.3 $\pm$ 4.0 & 95.6 $\pm$ 2.8\\
    \rowcolor{linecolor}\br\recon & \textbf{97.3 $\pm$ 1.9} & \textbf{98.9 $\pm$ 1.2} & \textbf{93.3 $\pm$ 3.9} & \textbf{95.8 $\pm$ 3.0} \\
    \midrule[0.6pt]
    \multicolumn{5}{c}{\textit{with Self-Supervised Representation Learning} ({\scshape Mlp-Linear})}\\
    \midrule[0.6pt]
    \br Point-MAE$^\dagger$ & 91.1 $\pm$ 5.6 & 91.7 $\pm$ 4.0 & 83.5 $\pm$ 6.1 & 89.7 $\pm$ 4.1\\
    \bv ACT & 91.8 $\pm$ 4.7 & 93.1 $\pm$ 4.2 & 84.5 $\pm$ 6.4 & 90.7 $\pm$ 4.3\\
    \rowcolor{linecolor}\br\recon & \textbf{96.9 $\pm$ 2.6} & \textbf{98.2 $\pm$ 1.4} & \textbf{93.6 $\pm$ 4.7} & \textbf{95.4 $\pm$ 2.6}\\
    \midrule[0.6pt]
    \multicolumn{5}{c}{\textit{with Self-Supervised Representation Learning} ({\scshape Mlp-$3$})}\\
    \midrule[0.6pt]
    \br Point-MAE$^\dagger$ & 95.0 $\pm$ 2.8 & 96.7 $\pm$ 2.4 & 90.6 $\pm$ 4.7 & 93.8 $\pm$ 5.0\\
    \bv ACT & 95.9 $\pm$ 2.2 & 97.7 $\pm$ 1.8 & 92.4 $\pm$ 5.0 & 94.7 $\pm$ 3.9\\
    \rowcolor{linecolor}\br\recon&\textbf{97.4 $\pm$ 2.2 }& \textbf{98.5 $\pm$ 1.4} & \textbf{93.6 $\pm$ 4.7}& \textbf{95.7 $\pm$ 2.7}\\
    \bottomrule[0.95pt]
    \end{tabular}
    }
\vspace{-10pt}
\end{table}
\end{center}

\vspace{-20pt}
\textbf{Implementation}~
We use the standard plain Transformer~\citep{AttentionIsAllYouNeed} built with 12 \recon-blocks with dimension 384 and a lightweight PointNet patch embedding~\citep{PointNet,PointNet++,PointBERT} as the 3D representation learner.
ShapeNet~\citep{ShapeNet15} is used to pretrain \recon, which contains $\sim$51K unique 3D CAD models covering 55 object categories. 
We follow~\citet{SCL20} for \textit{single-modal} contrastive modeling.
For \textit{cross-modal} setting, we utilize point clouds, RGB images, and free-form languages, where the limited 3D data is enlarged with rich texture and semantic knowledge within images and languages~\citep{CrossPoint22,ACT23}.
We obtain RGB images by rendering from 3D meshes and language descriptions by concatenating text prompts with category descriptions. 
We use by default an ImageNet~\citep{ImageNet09} pretrained Vision Transformer (ViT-B)~\citep{ViT} as the image view teacher, and we use the text encoder from CLIP~\citep{CLIP} as the text view teacher.
The image and text teacher encoders are frozen during pretraining, and Smooth $\ell_1$-based positive-only distillation~\citep{SimSiam,WhiteContrast21} is used.
The masked generative modeling follows~\citet{PointMAE}, where the reconstruction metric is $\ell_2$ Chamfer-Distance~\citep{ChamferDistance17}.
\cref{fig:framework} shows an overall illustration and more details can be found in Appendix~\ref{app:impl_detail}.
\vspace{-3pt}
\section{Experiments}
\vspace{-3pt}
\subsection{Transfer Learning on Downstream Tasks}
\vspace{-5pt}
\paragraph{Transfer Protocol}
We use the same classification heads and transfer learning protocols following~\citet{ACT23}: {\scshape Full}, {\scshape Mlp-Linear}, and {\scshape Mlp-$3$}. 

\textbf{3D Real-World Object Recognition}~
ScanObjectNN~\citep{ScanObjectNN19} is one of the most challenging 3D datasets, which covers $\sim$15K real-world objects from 15 categories.
For a fair comparison, we report the results with and without the voting strategy~\citep{RSCNN} separately. 
Note that we only use simple \textit{Rotation} as data augmentation in training following~\citet{ACT23}.
We report \textit{single-modal} (\recon-SMC) and \textit{cross-modal} (\recon-CMC, default if not otherwise specified) results on three model variants (see \cref{app:impl_detail}.) 
The results are shown in \cref{tab:scanobjectnn}, it is observed that: 
(i) With a comparable GFLOPS, the performance of our \recon-block (\textit{from scratch}) is improved by~2.5\% compared with that of standard Transformer under {\scshape Full} tuning protocol. 
Further, after the pre-training of \textit{reconstruction guided contrastive learning}, \recon\ gains a significant improvement of +11.3\% accuracy averaged on the three variant ScanObjectNN benchmarks.
(ii) Compared to other self-supervised learning (SSL) methods, our \recon\ achieves the best generalization across both single-modal and cross-modal on all transferring protocols. 
\eg, \recon\ outperforms Point-MAE by +5.6\% on three ScanObjectNN variants. 
(iii) Compared with any supervised or self-supervised method, our \recon\ achieves a new state-of-the-art that outperforms existing methods by a large margin. 
\begin{figure*}[h!]
    \begin{center}
    \includegraphics[width=0.8\linewidth]{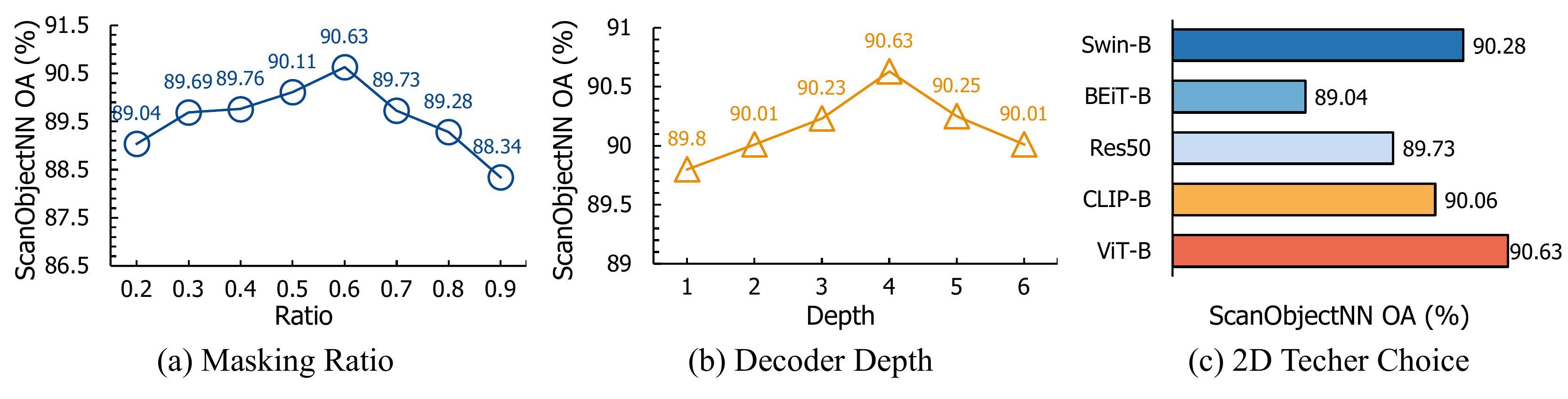}
    \vspace{-15pt}
    \caption{\textbf{Ablation study of masking ratio, decoder depth, and the 2D teacher encoder choice used during \recon~pretraining.} The masking ratio and decoder depth represent the ablation for the generative masked modeling stream, and the 2D teachers are used for contrastive cross-modal learning. }\label{fig:ablation}
    \end{center}
    \vspace{-22pt}
\end{figure*}

\textbf{3D Synthetic Object Recognition}~
ModelNet~\citep{ModelNet15} is one of the most classical datasets for synthetic 3D object recognition. It contains $\sim$12K meshed 3D CAD objects of 40 (ModelNet40) or 10 (ModelNet10) categories.
We conduct the evaluation on the ModelNet40 dataset, including fine-tuning and few-shot learning. 
We use \textit{Scale\&Translate} as data augmentation in training following~\citet{PointNet,PointNet++}. 
The results are shown in \cref{tab:scanobjectnn} and \cref{tab:few-shot}, respectively. 
It can be observed that our \recon\ achieves 94.7\% classification accuracy of ModelNet40 under {\scshape Full} protocol, improved by 2.7\% compared with the Transformer baseline. 
In the few-shot task, our \recon\ has also achieved the best performance under all protocols, especially under the {\scshape Mlp-$3$} and {\scshape Mlp-Linear} protocols.

\begin{table}[!t]
\caption{\textbf{Zero-shot 3D object classification domain transfer on ModelNet40 (MN-40) and ModelNet10 (MN-10)}. Top-1 accuracy (\%) is reported. Ensemb. denotes whether to use the ensemble strategy with multiple text inputs.} \label{tab:zeroshot}
\vspace{-8pt}
\begin{center}
\resizebox{\linewidth}{!}{
\begin{tabular}{lcccc}
\toprule[0.95pt]
Method  & Backbone & Ensemb. & MN-10 & MN-40\\
\midrule[0.6pt]
\bs PointCLIP~\cite{PointCLIP22} & ResNet-50 & $\times$ & 30.2 & 20.2 \\
\bv CLIP2Point~\cite{CLIP2Point22} & Transformer & $\checkmark$ & 66.6 & 49.4 \\
\rowcolor{linecolor1}\br\recon & Transformer & $\times$ & \textbf{74.2} & \textbf{60.6} \\
\rowcolor{linecolor}\br\recon & Transformer & $\checkmark$ & \textbf{75.6} & \textbf{61.7} \\
\bottomrule[0.95pt]
\end{tabular}
}
\end{center}
\vspace{-25pt}
\end{table}
\textbf{3D Zero-Shot Recognition}~
Similar to CLIP~\citep{CLIP}, our model aligns the feature space of languages and other modalities. 
Therefore, our model has a strong zero-shot capability. 
We use ModelNet~\citep{ModelNet15} dataset to conduct zero-shot evaluation, including ModelNet10 and ModelNet40. The results are shown in \cref{tab:zeroshot}.
Following PointCLIP~\citep{PointCLIP22}, We use prompt templates with the category label to generate text features. 
From \cref{tab:zeroshot}, it can be seen that our \recon\ surpasses all the zero-shot methods with CNN-based or Transformer-based backbones. 
Further, by using ensemble methods such as multi-prompt templates~\citep{CLIP2Point22}, our \recon\ achieves a Top-1 accuracy of 61.7\% on ModelNet40, which significantly outperforms PointCLIP and CLIP2Point by 41.5\% and 12.3\%, respectively. 

\begin{table}[!t]
\caption{\textbf{Ablation study on pretraining targets}. Overall accuracy (\%) without voting is reported.} \label{tab:target}
\begin{center}
\resizebox{0.9\linewidth}{!}{
\begin{tabular}{ccccc}
\toprule[0.95pt]
\multirow{2}{*}[-0.5ex]{Reconstruction} & \multicolumn{3}{c}{Contrastive} & \multirow{2}{*}[-0.5ex]{ScanObjectNN}\\
\cmidrule(lr){2-4} & text & image & self\\
\midrule[0.6pt]
$\times$ & $\checkmark$ & $\times$ & $\times$ & 85.60 \\
$\times$ & $\times$ & $\checkmark$ & $\times$ & 86.02 \\
$\times$ & $\times$ & $\times$ & $\checkmark$ & 85.57 \\
$\times$ & $\checkmark$ & $\checkmark$ & $\times$ & 86.49 \\
\midrule[0.6pt]
$\checkmark$ & $\times$ & $\times$ & $\times$ & 88.42 \\
$\checkmark$ & $\checkmark$ & $\times$ & $\times$ & 89.77 \\
$\checkmark$ & $\times$ & $\checkmark$ & $\times$ & 90.18 \\
$\checkmark$ & $\times$ & $\times$ & $\checkmark$ & 89.50 \\
$\checkmark$ & $\checkmark$ & $\checkmark$ & $\times$ & \cellcolor{linecolor}\textbf{90.63} \\
$\checkmark$ & $\checkmark$ & $\checkmark$ & $\checkmark$ & 89.98 \\
\bottomrule[0.95pt]
\end{tabular}
}
\vspace{-15pt}
\end{center}
\end{table}
\vspace{-3pt}
\subsection{Ablation Study}\label{sec:ablation}
\textbf{Hyper Parameter}~
In \cref{fig:ablation}, We show the ablation study on masking ratio, decoder depth, and the selection of the 2D image teacher during \recon\ pretraining. 
It can be observed that the optimal masking ratio and decoder depth is consistent with Point-MAE~\citep{PointMAE}, indicating that the pretraining model, which is friendly to downstream tasks, is also helpful for the guidance of contrastive learning.
The results also show that ViT~\citep{ViT}, as a 2D teacher, is superior to CLIP~\citep{CLIP}, Swin-Transformer (Swin)~\citep{SwinT}, ResNet~\citep{ResNet16} and BEiT~\citep{BEiT}. 
In addition, we find that CLIP, which is already aligned with languages, brings inferior performance to ViT.
This may be due to the reduction of diversity caused by the high similarity of features from the pre-aligned text teacher, which can be considered degenerating to only a single vision-language pretrained teacher.

\textbf{Pretraining Targets}~
To analyze the importance of the pretraining targets and verify the effect of reconstruction-guided contrastive modeling, we conduct an ablation study on the pretraining target. 
The results are shown in \cref{tab:target}.
It can be seen that: 
(i) When reconstruction guidance is not used, the performance of the contrastive model is very poor due to over-fitting on the limited 3D data.
(ii) The performance of single-modal contrastive learning is slightly weaker than that of cross-modal contrastive learning, and the improvements can not be shared.
Besides, we find that both 2D and text teachers can help improve performance without contradictions, and 2D teachers bring better improvement in learned representations generalization.

\begin{table}[!t]
\caption{\textbf{Ablation study on the contrastive metric}. Overall accuracy (\%) without voting is reported.} \label{tab:contrastive_loss}
\vspace{-5pt}
\begin{center}
\resizebox{0.9\linewidth}{!}{
\begin{tabular}{lcc}
\toprule[0.95pt]
Contrastive Metric & ScanObjectNN & ModelNet40\\
\midrule[0.6pt]
InfoNCE & 90.11 & 93.8\\
$\ell_2$ Distance & 89.64 & 93.6\\
Smooth $\ell_1$ & \cellcolor{linecolor}\textbf{90.63} & \cellcolor{linecolor}\textbf{94.1}\\
Cosine Similarity & 90.17 & 93.8\\
\bottomrule[0.95pt]
\end{tabular}
}
\vspace{-18pt}
\end{center}
\end{table}

\textbf{Contrastive Metric}~
\cref{tab:contrastive_loss} shows the ablation study on the contrastive metric.
Smooth $\ell_1$ distance achieves the best results in both tasks and is higher than the commonly used InfoNCE~\citep{InfoNCE}.
We argue that the reasons are two-fold:
(i) The cross-modal positive-only contrastive learning with the frozen teachers (stop-grad) has no risk of representation collapsing~\citep{SimSiam}, and it is not necessary to introduce negative samples.
(ii) ShapeNet dataset is full of household objects with limited semantic diversity, unlike ImageNet, which makes the negative samples noisy and confusing. 
These hard negatives are generally not easy for mining and may bring unsatisfactory optimization challenges~\citep{VSEPP}. 

\section{Discussions}\label{Discuss}
\subsection{What role does the reconstruction guidance play in contrastive learning?}
\vspace{-6pt}
To analyze the reason why \recon\ works, we record the contrastive loss on the \textit{test} split of ShapeNet (not used for pretraining) using cross-modal contrastive (CMC) and \textit{reconstruction guidance contrastive} (\recon-CMC), along with the corresponding fine-tuning accuracy on ScanObjectNN. 
\begin{figure}[t!]
    \begin{center}
    \includegraphics[width=0.95\linewidth]{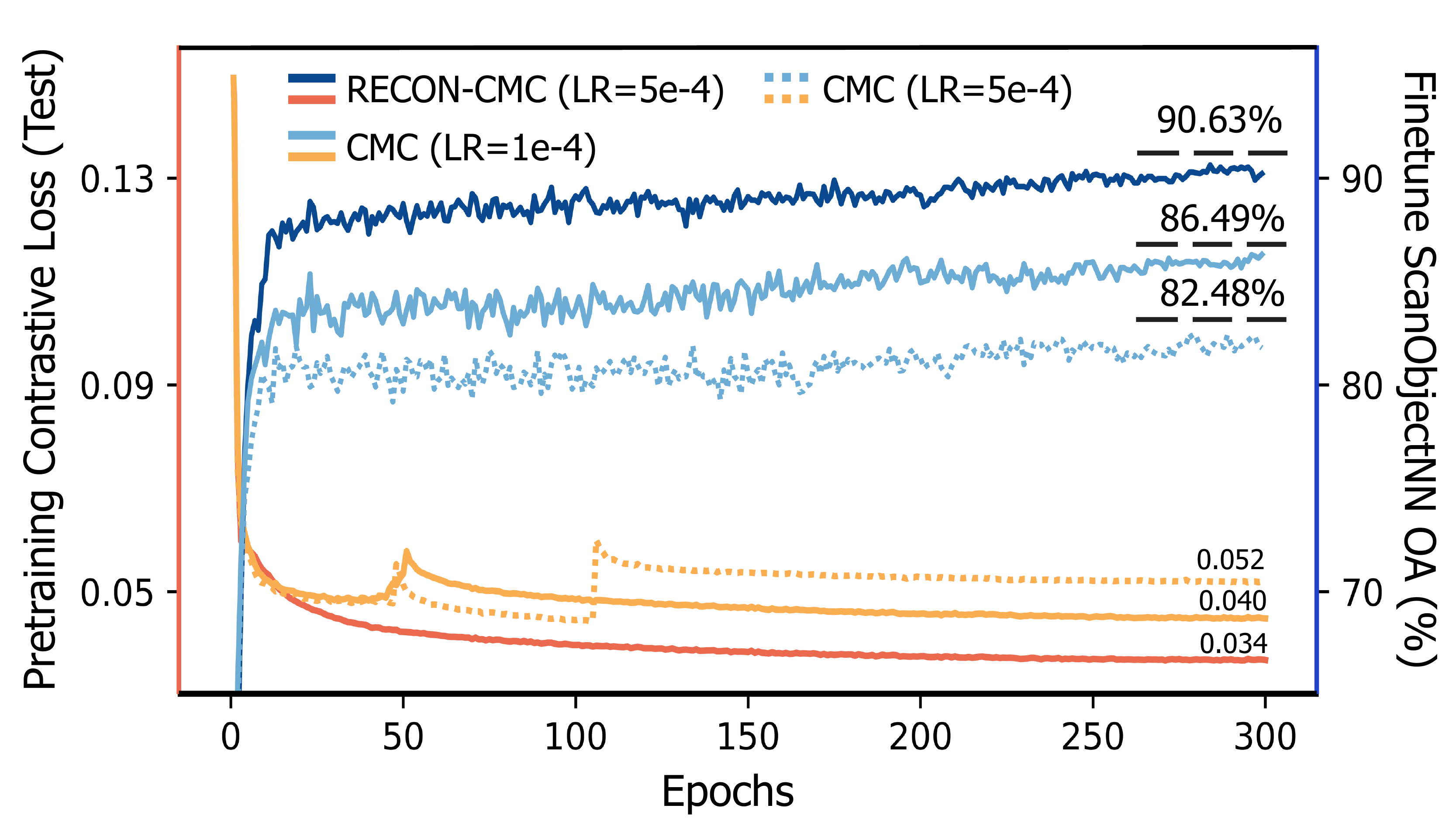}
    \vspace{-10pt}
    \caption{\textbf{Pretraining contrastive loss on ShapeNet \textit{test} split vs. finetuning accuracy (\%) on ScanObjectNN}. The pretraining \textit{test} loss (left) is plotted in orange, and the corresponding fine-tuning accuracy (right) is plotted in blue. LR: pretraining learning rate.
    }\label{fig:loss_curve}
    \vspace{-18pt}
    \end{center}
\end{figure}
The results are shown in Figure~\ref{fig:loss_curve}. 
It can be seen that the \textit{test} contrastive loss of our \recon-CMC is consistently lower than vanilla CMC, which converges to a lower value more stably (0.034 vs. 0.052), indicating that our \recon~brings superior generalization performance of the pretraining contrastive task. 
As a result, \recon-CMC learns to contrast without falling into shortcuts of trivial solutions, and the over-fitting issue during pretraining is alleviated.
The corresponding fine-tuning accuracy demonstrates this point, where a significantly superior generalization performance with better training efficiency of our \recon-CMC compared to vanilla CMC is achieved (90.63\% vs. 82.48\%).
Besides, we find that a lower learning rate improves vanilla CMC performance, demonstrating that contrastive models are prone to over-fit (see Appendix~\ref{app:add_ablation}).

\begin{table}[!t]
\caption{\textbf{Study of the \textit{stop-gradient} operation in \recon-block}. Overall accuracy (\%) without voting is reported.} \label{tab:detach}
\begin{center}
\resizebox{0.72\linewidth}{!}{
\begin{tabular}{ccc}
\toprule[0.95pt]
Stop-grad & ScanObjectNN & ModelNet40\\
\midrule[0.6pt]
$\times$ & 81.60 & 89.7\\
$\checkmark$ & \cellcolor{linecolor}\textbf{90.63} & \cellcolor{linecolor}\textbf{94.1}\\
\bottomrule[0.95pt]
\end{tabular}
}
\end{center}
\vspace{-15pt}
\end{table}

\vspace{-4pt}
\subsection{Can contrastive learning guide reconstruction?}
\vspace{-3pt}
As discussed in Sec.~\ref{sec:intro}, the learned patterns of contrastive and generative modeling are different, and we build \recon\ where the generative student guides the contrastive student.
What if we use global contrastive learning to guide the generative masked modeling?
To answer this question, we analyze the role of \textit{stop-gradient} in \recon-block.
The results are shown in Table~\ref{tab:detach}. 
It can be seen that \textit{without} stop-gradient, the performance of \recon~is seriously degraded (-9.03\% on ScanObjectNN and -4.4\% on ModelNet40).
We argue that the contrastive task can easily converge to a degenerated solution due to the limited 3D data, as discussed before, which in turn leads to noisy gradient and training signal to the generative modeling part during pretraining.
As a result, the guidance itself is disturbed and harmed, while
the model fails to learn representations with valid semantics.

\begin{figure}[t!]
    \begin{center}
    \includegraphics[width=\linewidth]{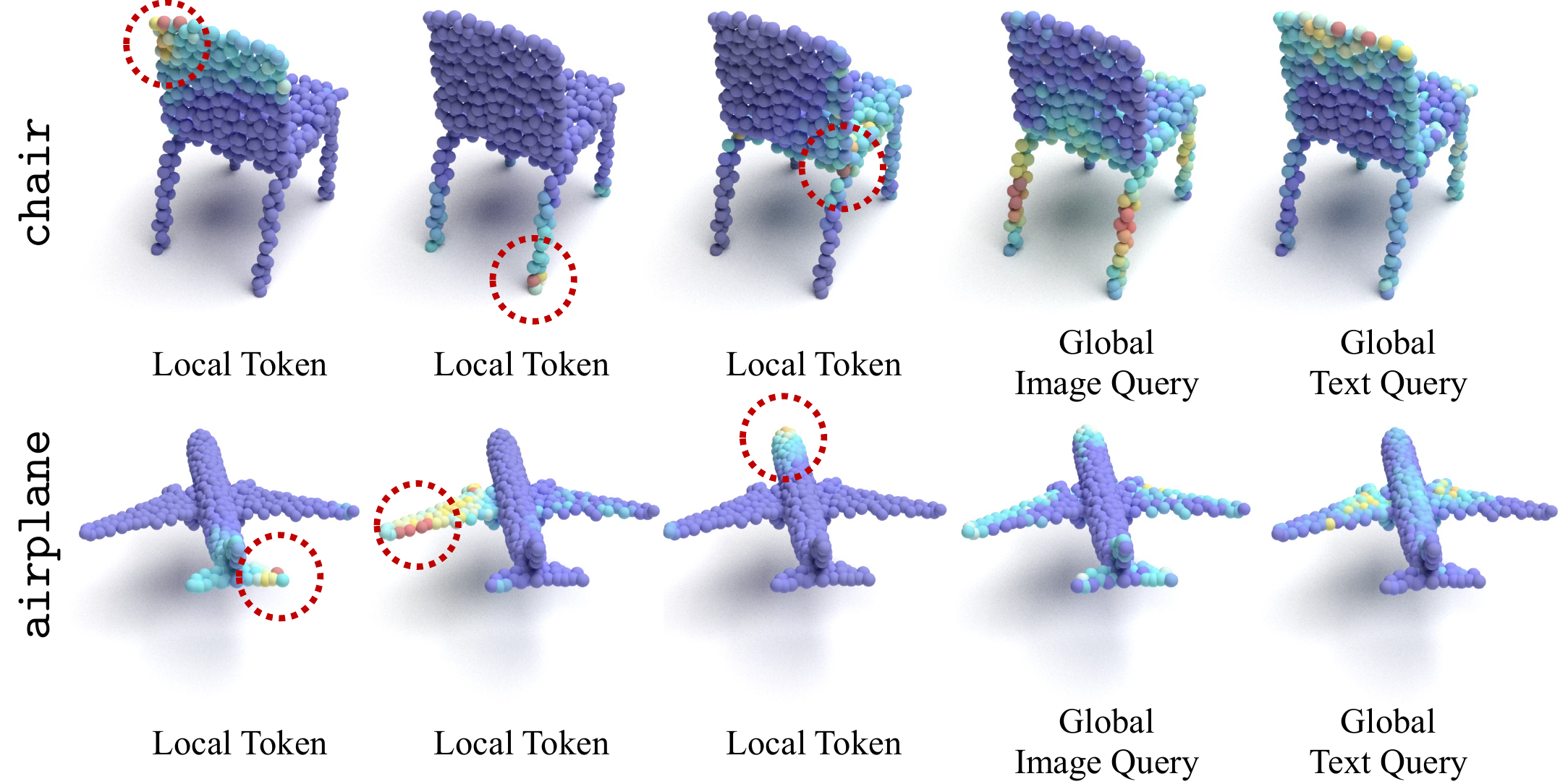}
    \vspace{-20pt}
    \caption{\textbf{Attention distribution visualization of local tokens and global queries learned by \recon}. We randomly select three local tokens, which are highlighted in red dashed circles.}\label{fig:atten_vis}
    \end{center}
    \vspace{-25pt}
\end{figure}
\vspace{-5pt}
\subsection{What is learned in \recon?}
\vspace{-4pt}
According to the design, \recon\ should have learned both locally focused and global 3D geometric understandings. 
In Figure~\ref{fig:atten_vis}, we visualize the attention distribution of randomly selected local tokens and global queries. 
The red and yellow parts are the key areas of attention, while the blue and purple parts are the areas of less attention.
It can be seen that the local tokens in 3D point clouds focus more on the geometric structure around the tokens themselves, while the global queries focus on the whole parts of the object. 
Interestingly, the local tokens may have learned some geometric understanding of symmetry.
For example, the token on the left wing of the \texttt{airplane} also noticed the right wing. 
In addition, we find that global image and text queries may have learned some complementary knowledge.
\vspace{-6pt}
\section{Conclusions}
\vspace{-3pt}
In this paper, we propose \textit{contrast with reconstruct} (\recon), which enjoys the merits of both generative masked modeling and contrastive modeling, while scalable to multimodal data to facilitate stronger 3D representation learning.
Our results show high-capacity data efficiency and generalizations on both pretraining and downstream representation transferring.
In particular, \recon\ achieves a new state-of-the-art on challenging real-world 3D object recognition.
By diving deeply into \recon, we emphasize the importance of reconstruction, which avoids the contrastive over-fitting issue due to limited 3D data.
Visualizations show that \recon\ indeed learns decoupled local and global representations in the proposed \recon-block.
\recon\ is a simple framework, and we hope more \recon-style models to be produced in the multi-modal learning community.


{
\small\bibliographystyle{icml2023}
\bibliography{main}

\begin{thebibliography}{94}
\providecommand{\natexlab}[1]{#1}
\providecommand{\url}[1]{\texttt{#1}}
\expandafter\ifx\csname urlstyle\endcsname\relax
  \providecommand{\doi}[1]{doi: #1}\else
  \providecommand{\doi}{doi: \begingroup \urlstyle{rm}\Url}\fi

\bibitem[Afham et~al.(2022)Afham, Dissanayake, Dissanayake, Dharmasiri,
  Thilakarathna, and Rodrigo]{CrossPoint22}
Afham, M., Dissanayake, I., Dissanayake, D., Dharmasiri, A., Thilakarathna, K.,
  and Rodrigo, R.
\newblock Crosspoint: Self-supervised cross-modal contrastive learning for 3d
  point cloud understanding.
\newblock In \emph{IEEE/CVF Conf. Comput. Vis. Pattern Recog. (CVPR)}, 2022.

\bibitem[Alayrac et~al.(2022)Alayrac, Donahue, Luc, Miech, Barr, Hasson, Lenc,
  Mensch, Millican, Reynolds, Ring, Rutherford, Cabi, Han, Gong, Samangooei,
  Monteiro, Menick, Borgeaud, Brock, Nematzadeh, Sharifzadeh, Binkowski,
  Barreira, Vinyals, Zisserman, and Simonyan]{Flamingo22}
Alayrac, J., Donahue, J., Luc, P., Miech, A., Barr, I., Hasson, Y., Lenc, K.,
  Mensch, A., Millican, K., Reynolds, M., Ring, R., Rutherford, E., Cabi, S.,
  Han, T., Gong, Z., Samangooei, S., Monteiro, M., Menick, J., Borgeaud, S.,
  Brock, A., Nematzadeh, A., Sharifzadeh, S., Binkowski, M., Barreira, R.,
  Vinyals, O., Zisserman, A., and Simonyan, K.
\newblock Flamingo: a visual language model for few-shot learning.
\newblock In \emph{Adv. Neural Inform. Process. Syst. (NeurIPS)}, 2022.

\bibitem[Bachman et~al.(2019)Bachman, Hjelm, and Buchwalter]{AugDeepInfoMax19}
Bachman, P., Hjelm, R.~D., and Buchwalter, W.
\newblock Learning representations by maximizing mutual information across
  views.
\newblock In \emph{Adv. Neural Inform. Process. Syst. (NeurIPS)}, 2019.

\bibitem[Bao et~al.(2022)Bao, Dong, Piao, and Wei]{BEiT}
Bao, H., Dong, L., Piao, S., and Wei, F.
\newblock Beit: {BERT} pre-training of image transformers.
\newblock In \emph{Int. Conf. Learn. Represent. (ICLR)}. OpenReview.net, 2022.

\bibitem[Belghazi et~al.(2018)Belghazi, Baratin, Rajeswar, Ozair, Bengio,
  Hjelm, and Courville]{MINE18}
Belghazi, M.~I., Baratin, A., Rajeswar, S., Ozair, S., Bengio, Y., Hjelm,
  R.~D., and Courville, A.~C.
\newblock Mutual information neural estimation.
\newblock In \emph{Proc. Int. Conf. Mach. Learn. (ICML)}, volume~80 of
  \emph{Proceedings of Machine Learning Research}, pp.\  530--539. {PMLR},
  2018.

\bibitem[Bommasani et~al.(2021)Bommasani, Hudson, Adeli, Altman, Arora, von
  Arx, Bernstein, Bohg, Bosselut, Brunskill, Brynjolfsson, Buch, Card,
  Castellon, Chatterji, Chen, Creel, Davis, Demszky, Donahue, Doumbouya,
  Durmus, Ermon, Etchemendy, Ethayarajh, Fei{-}Fei, Finn, Gale, Gillespie,
  Goel, Goodman, Grossman, Guha, Hashimoto, Henderson, Hewitt, Ho, Hong, Hsu,
  Huang, Icard, Jain, Jurafsky, Kalluri, Karamcheti, Keeling, Khani, Khattab,
  Koh, Krass, Krishna, Kuditipudi, and et~al.]{FoundationModel21}
Bommasani, R., Hudson, D.~A., Adeli, E., Altman, R., Arora, S., von Arx, S.,
  Bernstein, M.~S., Bohg, J., Bosselut, A., Brunskill, E., Brynjolfsson, E.,
  Buch, S., Card, D., Castellon, R., Chatterji, N.~S., Chen, A.~S., Creel, K.,
  Davis, J.~Q., Demszky, D., Donahue, C., Doumbouya, M., Durmus, E., Ermon, S.,
  Etchemendy, J., Ethayarajh, K., Fei{-}Fei, L., Finn, C., Gale, T., Gillespie,
  L., Goel, K., Goodman, N.~D., Grossman, S., Guha, N., Hashimoto, T.,
  Henderson, P., Hewitt, J., Ho, D.~E., Hong, J., Hsu, K., Huang, J., Icard,
  T., Jain, S., Jurafsky, D., Kalluri, P., Karamcheti, S., Keeling, G., Khani,
  F., Khattab, O., Koh, P.~W., Krass, M.~S., Krishna, R., Kuditipudi, R., and
  et~al.
\newblock On the opportunities and risks of foundation models.
\newblock \emph{CoRR}, abs/2108.07258, 2021.

\bibitem[Boser et~al.(1992)Boser, Guyon, and Vapnik]{MarginSVM92}
Boser, B.~E., Guyon, I., and Vapnik, V.
\newblock A training algorithm for optimal margin classifiers.
\newblock In \emph{ACM Conf. Comput. Learn. Theory (COLT)}, pp.\  144--152.
  {ACM}, 1992.

\bibitem[Brown et~al.(2020)Brown, Mann, Ryder, Subbiah, Kaplan, Dhariwal,
  Neelakantan, Shyam, Sastry, Askell, Agarwal, Herbert{-}Voss, Krueger,
  Henighan, Child, Ramesh, Ziegler, Wu, Winter, Hesse, Chen, Sigler, Litwin,
  Gray, Chess, Clark, Berner, McCandlish, Radford, Sutskever, and
  Amodei]{GPT3_20}
Brown, T.~B., Mann, B., Ryder, N., Subbiah, M., Kaplan, J., Dhariwal, P.,
  Neelakantan, A., Shyam, P., Sastry, G., Askell, A., Agarwal, S.,
  Herbert{-}Voss, A., Krueger, G., Henighan, T., Child, R., Ramesh, A.,
  Ziegler, D.~M., Wu, J., Winter, C., Hesse, C., Chen, M., Sigler, E., Litwin,
  M., Gray, S., Chess, B., Clark, J., Berner, C., McCandlish, S., Radford, A.,
  Sutskever, I., and Amodei, D.
\newblock Language models are few-shot learners.
\newblock In \emph{Adv. Neural Inform. Process. Syst. (NeurIPS)}, 2020.

\bibitem[Caron et~al.(2021)Caron, Touvron, Misra, J{\'{e}}gou, Mairal,
  Bojanowski, and Joulin]{DINO21}
Caron, M., Touvron, H., Misra, I., J{\'{e}}gou, H., Mairal, J., Bojanowski, P.,
  and Joulin, A.
\newblock Emerging properties in self-supervised vision transformers.
\newblock In \emph{Int. Conf. Comput. Vis. (ICCV)}, pp.\  9630--9640. {IEEE},
  2021.

\bibitem[Chang et~al.(2015)Chang, Funkhouser, Guibas, Hanrahan, Huang, Li,
  Savarese, Savva, Song, Su, Xiao, Yi, and Yu]{ShapeNet15}
Chang, A.~X., Funkhouser, T.~A., Guibas, L.~J., Hanrahan, P., Huang, Q., Li,
  Z., Savarese, S., Savva, M., Song, S., Su, H., Xiao, J., Yi, L., and Yu, F.
\newblock Shapenet: An information-rich 3d model repository.
\newblock \emph{CoRR}, abs/1512.03012, 2015.

\bibitem[Chen et~al.(2020)Chen, Kornblith, Norouzi, and Hinton]{SimCLR}
Chen, T., Kornblith, S., Norouzi, M., and Hinton, G.~E.
\newblock A simple framework for contrastive learning of visual
  representations.
\newblock In \emph{Proc. Int. Conf. Mach. Learn. (ICML)}, volume 119 of
  \emph{Proceedings of Machine Learning Research}, pp.\  1597--1607. {PMLR},
  2020.

\bibitem[Chen \& He(2021)Chen and He]{SimSiam}
Chen, X. and He, K.
\newblock Exploring simple siamese representation learning.
\newblock In \emph{IEEE/CVF Conf. Comput. Vis. Pattern Recog. (CVPR)}, pp.\
  15750--15758, 2021.

\bibitem[Chen et~al.(2021)Chen, Xie, and He]{MoCoThree21}
Chen, X., Xie, S., and He, K.
\newblock An empirical study of training self-supervised vision transformers.
\newblock In \emph{Int. Conf. Comput. Vis. (ICCV)}, pp.\  9620--9629. {IEEE},
  2021.

\bibitem[Chen et~al.(2022)Chen, Nie{\ss}ner, and Dai]{4DContrast22}
Chen, Y., Nie{\ss}ner, M., and Dai, A.
\newblock 4dcontrast: Contrastive learning with dynamic correspondences for 3d
  scene understanding.
\newblock In \emph{Eur. Conf. Comput. Vis. (ECCV)}, 2022.

\bibitem[Dangovski et~al.(2022)Dangovski, Jing, Loh, Han, Srivastava, Cheung,
  Agrawal, and Soljacic]{EquivariantSSL22}
Dangovski, R., Jing, L., Loh, C., Han, S., Srivastava, A., Cheung, B., Agrawal,
  P., and Soljacic, M.
\newblock Equivariant self-supervised learning: Encouraging equivariance in
  representations.
\newblock In \emph{Int. Conf. Learn. Represent. (ICLR)}. OpenReview.net, 2022.

\bibitem[Deng et~al.(2009)Deng, Dong, Socher, Li, Li, and
  Fei{-}Fei]{ImageNet09}
Deng, J., Dong, W., Socher, R., Li, L., Li, K., and Fei{-}Fei, L.
\newblock Imagenet: {A} large-scale hierarchical image database.
\newblock In \emph{IEEE/CVF Conf. Comput. Vis. Pattern Recog. (CVPR)}, 2009.

\bibitem[Devlin et~al.(2019)Devlin, Chang, Lee, and Toutanova]{BERT}
Devlin, J., Chang, M., Lee, K., and Toutanova, K.
\newblock {BERT:} pre-training of deep bidirectional transformers for language
  understanding.
\newblock In \emph{Proceedings of the 2019 Conference of the North American
  Chapter of the Association for Computational Linguistics: Human Language
  Technologies, {NAACL-HLT} 2019, Minneapolis, MN, USA, June 2-7, 2019, Volume
  1 (Long and Short Papers)}, pp.\  4171--4186. Association for Computational
  Linguistics, 2019.

\bibitem[Dong et~al.(2023)Dong, Qi, Zhang, Zhang, Sun, Ge, Yi, and Ma]{ACT23}
Dong, R., Qi, Z., Zhang, L., Zhang, J., Sun, J., Ge, Z., Yi, L., and Ma, K.
\newblock Autoencoders as cross-modal teachers: Can pretrained 2d image
  transformers help 3d representation learning?
\newblock In \emph{Int. Conf. Learn. Represent. (ICLR)}, 2023.

\bibitem[Dong et~al.(2022)Dong, Bao, Zhang, Chen, Gu, Zhang, Yuan, Chen, Wen,
  and Yu]{TuneCLIP22}
Dong, X., Bao, J., Zhang, T., Chen, D., Gu, S., Zhang, W., Yuan, L., Chen, D.,
  Wen, F., and Yu, N.
\newblock {CLIP} itself is a strong fine-tuner: Achieving 85.7{\%} and 88.0{\%}
  top-1 accuracy with vit-b and vit-l on imagenet.
\newblock \emph{CoRR}, abs/2212.06138, 2022.

\bibitem[Dosovitskiy et~al.(2021)Dosovitskiy, Beyer, Kolesnikov, Weissenborn,
  Zhai, Unterthiner, Dehghani, Minderer, Heigold, Gelly, Uszkoreit, and
  Houlsby]{ViT}
Dosovitskiy, A., Beyer, L., Kolesnikov, A., Weissenborn, D., Zhai, X.,
  Unterthiner, T., Dehghani, M., Minderer, M., Heigold, G., Gelly, S.,
  Uszkoreit, J., and Houlsby, N.
\newblock An image is worth 16x16 words: Transformers for image recognition at
  scale.
\newblock In \emph{Int. Conf. Learn. Represent. (ICLR)}, 2021.

\bibitem[Ermolov et~al.(2021)Ermolov, Siarohin, Sangineto, and
  Sebe]{WhiteContrast21}
Ermolov, A., Siarohin, A., Sangineto, E., and Sebe, N.
\newblock Whitening for self-supervised representation learning.
\newblock In \emph{Proc. Int. Conf. Mach. Learn. (ICML)}, volume 139 of
  \emph{Proceedings of Machine Learning Research}, pp.\  3015--3024. {PMLR},
  2021.

\bibitem[Faghri et~al.(2018)Faghri, Fleet, Kiros, and Fidler]{VSEPP}
Faghri, F., Fleet, D.~J., Kiros, J.~R., and Fidler, S.
\newblock {VSE++:} improving visual-semantic embeddings with hard negatives.
\newblock In \emph{Brit. Mach. Vis. Conf. (BMVC)}, pp.\ ~12. {BMVA} Press,
  2018.

\bibitem[Fan et~al.(2017)Fan, Su, and Guibas]{ChamferDistance17}
Fan, H., Su, H., and Guibas, L.~J.
\newblock A point set generation network for 3d object reconstruction from a
  single image.
\newblock In \emph{IEEE/CVF Conf. Comput. Vis. Pattern Recog. (CVPR)}, 2017.

\bibitem[Goyal et~al.(2021)Goyal, Law, Liu, Newell, and Deng]{SimpleView}
Goyal, A., Law, H., Liu, B., Newell, A., and Deng, J.
\newblock Revisiting point cloud shape classification with a simple and
  effective baseline.
\newblock In \emph{Proc. Int. Conf. Mach. Learn. (ICML)}, volume 139 of
  \emph{Proceedings of Machine Learning Research}, pp.\  3809--3820. {PMLR},
  2021.

\bibitem[Goyal et~al.(2019)Goyal, Mahajan, Gupta, and Misra]{BenchmarkSSL19}
Goyal, P., Mahajan, D., Gupta, A., and Misra, I.
\newblock Scaling and benchmarking self-supervised visual representation
  learning.
\newblock In \emph{Int. Conf. Comput. Vis. (ICCV)}, pp.\  6390--6399. {IEEE},
  2019.

\bibitem[Grill et~al.(2020)Grill, Strub, Altch{\'{e}}, Tallec, Richemond,
  Buchatskaya, Doersch, Pires, Guo, Azar, Piot, Kavukcuoglu, Munos, and
  Valko]{BYOL}
Grill, J., Strub, F., Altch{\'{e}}, F., Tallec, C., Richemond, P.~H.,
  Buchatskaya, E., Doersch, C., Pires, B.~{\'{A}}., Guo, Z., Azar, M.~G., Piot,
  B., Kavukcuoglu, K., Munos, R., and Valko, M.
\newblock Bootstrap your own latent - {A} new approach to self-supervised
  learning.
\newblock In \emph{Adv. Neural Inform. Process. Syst. (NeurIPS)}, 2020.

\bibitem[Guo et~al.(2021)Guo, Cai, Liu, Mu, Martin, and Hu]{PCT}
Guo, M., Cai, J., Liu, Z., Mu, T., Martin, R.~R., and Hu, S.
\newblock {PCT:} point cloud transformer.
\newblock \emph{Comput. Vis. Media}, 7\penalty0 (2):\penalty0 187--199, 2021.

\bibitem[Hadsell et~al.(2006)Hadsell, Chopra, and LeCun]{LeCunContrastive}
Hadsell, R., Chopra, S., and LeCun, Y.
\newblock Dimensionality reduction by learning an invariant mapping.
\newblock In \emph{IEEE/CVF Conf. Comput. Vis. Pattern Recog. (CVPR)}, pp.\
  1735--1742, 2006.

\bibitem[Hamdi et~al.(2021)Hamdi, Giancola, and Ghanem]{MVTN}
Hamdi, A., Giancola, S., and Ghanem, B.
\newblock {MVTN:} multi-view transformation network for 3d shape recognition.
\newblock In \emph{Int. Conf. Comput. Vis. (ICCV)}, pp.\  1--11. {IEEE}, 2021.

\bibitem[He et~al.(2016)He, Zhang, Ren, and Sun]{ResNet16}
He, K., Zhang, X., Ren, S., and Sun, J.
\newblock Deep residual learning for image recognition.
\newblock In \emph{IEEE/CVF Conf. Comput. Vis. Pattern Recog. (CVPR)}, pp.\
  770--778. {IEEE} Computer Society, 2016.

\bibitem[He et~al.(2020)He, Fan, Wu, Xie, and Girshick]{MoCo}
He, K., Fan, H., Wu, Y., Xie, S., and Girshick, R.~B.
\newblock Momentum contrast for unsupervised visual representation learning.
\newblock In \emph{IEEE/CVF Conf. Comput. Vis. Pattern Recog. (CVPR)}, pp.\
  9726--9735. Computer Vision Foundation / {IEEE}, 2020.

\bibitem[He et~al.(2022)He, Chen, Xie, Li, Doll{\'{a}}r, and Girshick]{MAE}
He, K., Chen, X., Xie, S., Li, Y., Doll{\'{a}}r, P., and Girshick, R.~B.
\newblock Masked autoencoders are scalable vision learners.
\newblock In \emph{IEEE/CVF Conf. Comput. Vis. Pattern Recog. (CVPR)}, 2022.

\bibitem[Hinton et~al.(2015)Hinton, Vinyals, and Dean]{HintonKD15}
Hinton, G.~E., Vinyals, O., and Dean, J.
\newblock Distilling the knowledge in a neural network.
\newblock In \emph{Adv. Neural Inform. Process. Syst. (NeurIPS)}, volume
  abs/1503.02531, 2015.

\bibitem[Hjelm et~al.(2019)Hjelm, Fedorov, Lavoie{-}Marchildon, Grewal,
  Bachman, Trischler, and Bengio]{DeepInfoMax19}
Hjelm, R.~D., Fedorov, A., Lavoie{-}Marchildon, S., Grewal, K., Bachman, P.,
  Trischler, A., and Bengio, Y.
\newblock Learning deep representations by mutual information estimation and
  maximization.
\newblock In \emph{Int. Conf. Learn. Represent. (ICLR)}, 2019.

\bibitem[Huang et~al.(2022)Huang, Dong, Yang, Huang, Lau, Ouyang, and
  Zuo]{CLIP2Point22}
Huang, T., Dong, B., Yang, Y., Huang, X., Lau, R. W.~H., Ouyang, W., and Zuo,
  W.
\newblock Clip2point: Transfer {CLIP} to point cloud classification with
  image-depth pre-training.
\newblock \emph{CoRR}, abs/2210.01055, 2022.

\bibitem[Jing et~al.(2021)Jing, Vahdani, Tan, and Tian]{CenterLoss21}
Jing, L., Vahdani, E., Tan, J., and Tian, Y.
\newblock Cross-modal center loss for 3d cross-modal retrieval.
\newblock In \emph{IEEE/CVF Conf. Comput. Vis. Pattern Recog. (CVPR)}, pp.\
  3142--3151. Computer Vision Foundation / {IEEE}, 2021.

\bibitem[Khosla et~al.(2020)Khosla, Teterwak, Wang, Sarna, Tian, Isola,
  Maschinot, Liu, and Krishnan]{SCL20}
Khosla, P., Teterwak, P., Wang, C., Sarna, A., Tian, Y., Isola, P., Maschinot,
  A., Liu, C., and Krishnan, D.
\newblock Supervised contrastive learning.
\newblock In \emph{Adv. Neural Inform. Process. Syst. (NeurIPS)}, 2020.

\bibitem[Kong \& Zhang(2023)Kong and Zhang]{InvariantMIM22}
Kong, X. and Zhang, X.
\newblock Understanding masked image modeling via learning occlusion invariant
  feature.
\newblock In \emph{IEEE/CVF Conf. Comput. Vis. Pattern Recog. (CVPR)}, 2023.

\bibitem[Kosmann-Schwarzbach(2011)]{InvarianceTheory11}
Kosmann-Schwarzbach, Y.
\newblock \emph{The Noether Theorems}, pp.\  55--64.
\newblock Springer New York, New York, NY, 2011.
\newblock ISBN 978-0-387-87868-3.
\newblock \doi{10.1007/978-0-387-87868-3_3}.
\newblock URL \url{https://doi.org/10.1007/978-0-387-87868-3_3}.

\bibitem[Li et~al.(2021)Li, Selvaraju, Gotmare, Joty, Xiong, and Hoi]{albef21}
Li, J., Selvaraju, R.~R., Gotmare, A., Joty, S.~R., Xiong, C., and Hoi, S.~C.
\newblock Align before fuse: Vision and language representation learning with
  momentum distillation.
\newblock In \emph{Adv. Neural Inform. Process. Syst. (NeurIPS)}, pp.\
  9694--9705, 2021.

\bibitem[Li et~al.(2023)Li, Li, Savarese, and Hoi]{BLIP223}
Li, J., Li, D., Savarese, S., and Hoi, S. C.~H.
\newblock {BLIP-2:} bootstrapping language-image pre-training with frozen image
  encoders and large language models.
\newblock \emph{CoRR}, abs/2301.12597, 2023.

\bibitem[Li et~al.(2022{\natexlab{a}})Li, Zhang, Zhang, Yang, Li, Zhong, Wang,
  Yuan, Zhang, Hwang, Chang, and Gao]{GLIP22}
Li, L.~H., Zhang, P., Zhang, H., Yang, J., Li, C., Zhong, Y., Wang, L., Yuan,
  L., Zhang, L., Hwang, J., Chang, K., and Gao, J.
\newblock Grounded language-image pre-training.
\newblock In \emph{IEEE/CVF Conf. Comput. Vis. Pattern Recog. (CVPR)}, pp.\
  10955--10965. {IEEE}, 2022{\natexlab{a}}.

\bibitem[Li et~al.(2018)Li, Bu, Sun, Wu, Di, and Chen]{PointCNN}
Li, Y., Bu, R., Sun, M., Wu, W., Di, X., and Chen, B.
\newblock Pointcnn: Convolution on x-transformed points.
\newblock In \emph{Adv. Neural Inform. Process. Syst. (NeurIPS)}, pp.\
  828--838, 2018.

\bibitem[Li et~al.(2022{\natexlab{b}})Li, Fan, Hu, Feichtenhofer, and
  He]{FLIP22}
Li, Y., Fan, H., Hu, R., Feichtenhofer, C., and He, K.
\newblock Scaling language-image pre-training via masking.
\newblock \emph{CoRR}, abs/2212.00794, 2022{\natexlab{b}}.

\bibitem[Liu et~al.(2022)Liu, Cai, and Lee]{MaskPoint}
Liu, H., Cai, M., and Lee, Y.~J.
\newblock Masked discrimination for self-supervised learning on point clouds.
\newblock In \emph{Eur. Conf. Comput. Vis. (ECCV)}, 2022.

\bibitem[Liu et~al.(2019)Liu, Fan, Xiang, and Pan]{RSCNN}
Liu, Y., Fan, B., Xiang, S., and Pan, C.
\newblock Relation-shape convolutional neural network for point cloud analysis.
\newblock In \emph{IEEE/CVF Conf. Comput. Vis. Pattern Recog. (CVPR)}, pp.\
  8895--8904. Computer Vision Foundation / {IEEE}, 2019.

\bibitem[Liu et~al.(2021{\natexlab{a}})Liu, Fan, Zhang, Dong, Funkhouser, and
  Yi]{TupleInfoNCE21}
Liu, Y., Fan, Q., Zhang, S., Dong, H., Funkhouser, T.~A., and Yi, L.
\newblock Contrastive multimodal fusion with tupleinfonce.
\newblock In \emph{Int. Conf. Comput. Vis. (ICCV)}, pp.\  734--743. {IEEE},
  2021{\natexlab{a}}.

\bibitem[Liu et~al.(2021{\natexlab{b}})Liu, Lin, Cao, Hu, Wei, Zhang, Lin, and
  Guo]{SwinT}
Liu, Z., Lin, Y., Cao, Y., Hu, H., Wei, Y., Zhang, Z., Lin, S., and Guo, B.
\newblock Swin transformer: Hierarchical vision transformer using shifted
  windows.
\newblock In \emph{Int. Conf. Comput. Vis. (ICCV)}, pp.\  9992--10002. {IEEE},
  2021{\natexlab{b}}.

\bibitem[Loshchilov \& Hutter(2017)Loshchilov and Hutter]{CosineLRSGDR}
Loshchilov, I. and Hutter, F.
\newblock {SGDR:} stochastic gradient descent with warm restarts.
\newblock In \emph{Int. Conf. Learn. Represent. (ICLR)}. OpenReview.net, 2017.

\bibitem[Loshchilov \& Hutter(2019)Loshchilov and Hutter]{AdamW19}
Loshchilov, I. and Hutter, F.
\newblock Decoupled weight decay regularization.
\newblock In \emph{Int. Conf. Learn. Represent. (ICLR)}, 2019.

\bibitem[Ma et~al.(2022)Ma, Qin, You, Ran, and Fu]{PointMLP}
Ma, X., Qin, C., You, H., Ran, H., and Fu, Y.
\newblock Rethinking network design and local geometry in point cloud: {A}
  simple residual {MLP} framework.
\newblock In \emph{Int. Conf. Learn. Represent. (ICLR)}. OpenReview.net, 2022.

\bibitem[OpenAI(2022)]{ChatGPT}
OpenAI.
\newblock Introducing chatgpt.
\newblock 2022.
\newblock URL \url{https://openai.com/blog/chatgpt}.

\bibitem[Ouyang et~al.(2022)Ouyang, Wu, Jiang, Almeida, Wainwright, Mishkin,
  Zhang, Agarwal, Slama, Ray, Schulman, Hilton, Kelton, Miller, Simens, Askell,
  Welinder, Christiano, Leike, and Lowe]{InstructGPT22}
Ouyang, L., Wu, J., Jiang, X., Almeida, D., Wainwright, C.~L., Mishkin, P.,
  Zhang, C., Agarwal, S., Slama, K., Ray, A., Schulman, J., Hilton, J., Kelton,
  F., Miller, L., Simens, M., Askell, A., Welinder, P., Christiano, P.~F.,
  Leike, J., and Lowe, R.
\newblock Training language models to follow instructions with human feedback.
\newblock \emph{CoRR}, abs/2203.02155, 2022.

\bibitem[Pang et~al.(2022)Pang, Wang, Tay, Liu, Tian, and Yuan]{PointMAE}
Pang, Y., Wang, W., Tay, F. E.~H., Liu, W., Tian, Y., and Yuan, L.
\newblock Masked autoencoders for point cloud self-supervised learning.
\newblock In \emph{Eur. Conf. Comput. Vis. (ECCV)}, 2022.

\bibitem[Park \& Kwak(2020)Park and Kwak]{Feed20}
Park, S. and Kwak, N.
\newblock Feature-level ensemble knowledge distillation for aggregating
  knowledge from multiple networks.
\newblock In Giacomo, G.~D., Catal{\'{a}}, A., Dilkina, B., Milano, M., Barro,
  S., Bugar{\'{\i}}n, A., and Lang, J. (eds.), \emph{Eur. Conf. Artif. Intell.
  (ECAI)}, volume 325 of \emph{Frontiers in Artificial Intelligence and
  Applications}, pp.\  1411--1418. {IOS} Press, 2020.

\bibitem[Qi et~al.(2017{\natexlab{a}})Qi, Su, Mo, and Guibas]{PointNet}
Qi, C.~R., Su, H., Mo, K., and Guibas, L.~J.
\newblock Pointnet: Deep learning on point sets for 3d classification and
  segmentation.
\newblock In \emph{IEEE/CVF Conf. Comput. Vis. Pattern Recog. (CVPR)}, pp.\
  77--85, 2017{\natexlab{a}}.

\bibitem[Qi et~al.(2017{\natexlab{b}})Qi, Yi, Su, and Guibas]{PointNet++}
Qi, C.~R., Yi, L., Su, H., and Guibas, L.~J.
\newblock Pointnet++: Deep hierarchical feature learning on point sets in a
  metric space.
\newblock In \emph{Adv. Neural Inform. Process. Syst. (NIPS)}, pp.\
  5099--5108, 2017{\natexlab{b}}.

\bibitem[Qian et~al.(2022)Qian, Li, Peng, Mai, Hammoud, Elhoseiny, and
  Ghanem]{PointNext}
Qian, G., Li, Y., Peng, H., Mai, J., Hammoud, H. A. A.~K., Elhoseiny, M., and
  Ghanem, B.
\newblock Pointnext: Revisiting pointnet++ with improved training and scaling
  strategies.
\newblock In \emph{Adv. Neural Inform. Process. Syst. (NeurIPS)}, 2022.

\bibitem[Radford et~al.(2018)Radford, Narasimhan, Salimans, Sutskever,
  et~al.]{GPTv1_18}
Radford, A., Narasimhan, K., Salimans, T., Sutskever, I., et~al.
\newblock Improving language understanding by generative pre-training.
\newblock 2018.

\bibitem[Radford et~al.(2021)Radford, Kim, Hallacy, Ramesh, Goh, Agarwal,
  Sastry, Askell, Mishkin, Clark, Krueger, and Sutskever]{CLIP}
Radford, A., Kim, J.~W., Hallacy, C., Ramesh, A., Goh, G., Agarwal, S., Sastry,
  G., Askell, A., Mishkin, P., Clark, J., Krueger, G., and Sutskever, I.
\newblock Learning transferable visual models from natural language
  supervision.
\newblock In \emph{Proc. Int. Conf. Mach. Learn. (ICML)}, volume 139 of
  \emph{Proceedings of Machine Learning Research}, pp.\  8748--8763. {PMLR},
  2021.

\bibitem[Ramesh et~al.(2021)Ramesh, Pavlov, Goh, Gray, Voss, Radford, Chen, and
  Sutskever]{DALL-E}
Ramesh, A., Pavlov, M., Goh, G., Gray, S., Voss, C., Radford, A., Chen, M., and
  Sutskever, I.
\newblock Zero-shot text-to-image generation.
\newblock In \emph{Proc. Int. Conf. Mach. Learn. (ICML)}, volume 139 of
  \emph{Proceedings of Machine Learning Research}, pp.\  8821--8831. {PMLR},
  2021.

\bibitem[Rombach et~al.(2022)Rombach, Blattmann, Lorenz, Esser, and
  Ommer]{StableDiffusion22}
Rombach, R., Blattmann, A., Lorenz, D., Esser, P., and Ommer, B.
\newblock High-resolution image synthesis with latent diffusion models.
\newblock In \emph{IEEE/CVF Conf. Comput. Vis. Pattern Recog. (CVPR)}, pp.\
  10674--10685. {IEEE}, 2022.

\bibitem[Ruder(2017)]{MTLOverview}
Ruder, S.
\newblock An overview of multi-task learning in deep neural networks.
\newblock \emph{CoRR}, abs/1706.05098, 2017.

\bibitem[Schuhmann et~al.(2022)Schuhmann, Beaumont, Vencu, Gordon, Wightman,
  Cherti, Coombes, Katta, Mullis, Wortsman, Schramowski, Kundurthy, Crowson,
  Schmidt, Kaczmarczyk, and Jitsev]{LAION5B2022}
Schuhmann, C., Beaumont, R., Vencu, R., Gordon, C., Wightman, R., Cherti, M.,
  Coombes, T., Katta, A., Mullis, C., Wortsman, M., Schramowski, P., Kundurthy,
  S., Crowson, K., Schmidt, L., Kaczmarczyk, R., and Jitsev, J.
\newblock {LAION-5B:} an open large-scale dataset for training next generation
  image-text models.
\newblock \emph{CoRR}, abs/2210.08402, 2022.

\bibitem[Tao et~al.(2022)Tao, Zhu, Huang, Qiao, Wang, and Dai]{sim22}
Tao, C., Zhu, X., Huang, G., Qiao, Y., Wang, X., and Dai, J.
\newblock Siamese image modeling for self-supervised vision representation
  learning.
\newblock \emph{CoRR}, abs/2206.01204, 2022.

\bibitem[Tian et~al.(2020{\natexlab{a}})Tian, Krishnan, and Isola]{CMC20}
Tian, Y., Krishnan, D., and Isola, P.
\newblock Contrastive multiview coding.
\newblock In Vedaldi, A., Bischof, H., Brox, T., and Frahm, J. (eds.),
  \emph{Eur. Conf. Comput. Vis. (ECCV)}, volume 12356 of \emph{Lecture Notes in
  Computer Science}, pp.\  776--794. Springer, 2020{\natexlab{a}}.

\bibitem[Tian et~al.(2020{\natexlab{b}})Tian, Krishnan, and Isola]{CRD20}
Tian, Y., Krishnan, D., and Isola, P.
\newblock Contrastive representation distillation.
\newblock In \emph{Int. Conf. Learn. Represent. (ICLR)}, 2020{\natexlab{b}}.

\bibitem[Tian et~al.(2020{\natexlab{c}})Tian, Sun, Poole, Krishnan, Schmid, and
  Isola]{GoodViewContrast20}
Tian, Y., Sun, C., Poole, B., Krishnan, D., Schmid, C., and Isola, P.
\newblock What makes for good views for contrastive learning?
\newblock In \emph{Adv. Neural Inform. Process. Syst. (NeurIPS)},
  2020{\natexlab{c}}.

\bibitem[Uy et~al.(2019)Uy, Pham, Hua, Nguyen, and Yeung]{ScanObjectNN19}
Uy, M.~A., Pham, Q.-H., Hua, B.-S., Nguyen, T., and Yeung, S.-K.
\newblock Revisiting point cloud classification: A new benchmark dataset and
  classification model on real-world data.
\newblock In \emph{IEEE/CVF Conf. Comput. Vis. Pattern Recog. (CVPR)}, pp.\
  1588--1597, 2019.

\bibitem[van~den Oord et~al.(2018)van~den Oord, Li, and Vinyals]{InfoNCE}
van~den Oord, A., Li, Y., and Vinyals, O.
\newblock Representation learning with contrastive predictive coding.
\newblock \emph{CoRR}, abs/1807.03748, 2018.

\bibitem[Vapnik(1998)]{SLTheory98}
Vapnik, V.
\newblock \emph{Statistical learning theory}.
\newblock Wiley, 1998.
\newblock ISBN 978-0-471-03003-4.

\bibitem[Vaswani et~al.(2017)Vaswani, Shazeer, Parmar, Uszkoreit, Jones, Gomez,
  Kaiser, and Polosukhin]{AttentionIsAllYouNeed}
Vaswani, A., Shazeer, N., Parmar, N., Uszkoreit, J., Jones, L., Gomez, A.~N.,
  Kaiser, L., and Polosukhin, I.
\newblock Attention is all you need.
\newblock In \emph{Adv. Neural Inform. Process. Syst. (NIPS)}, pp.\
  5998--6008, 2017.

\bibitem[Wang et~al.(2021)Wang, Liu, Yue, Lasenby, and Kusner]{OcCo}
Wang, H., Liu, Q., Yue, X., Lasenby, J., and Kusner, M.~J.
\newblock Unsupervised point cloud pre-training via occlusion completion.
\newblock In \emph{Int. Conf. Comput. Vis. (ICCV)}, pp.\  9782--9792, 2021.

\bibitem[Wang \& Yoon(2022)Wang and Yoon]{KDOverview22}
Wang, L. and Yoon, K.
\newblock Knowledge distillation and student-teacher learning for visual
  intelligence: {A} review and new outlooks.
\newblock \emph{IEEE Trans. Pattern Anal. Mach. Intell. (TPAMI)}, 44\penalty0
  (6):\penalty0 3048--3068, 2022.

\bibitem[Wang \& Qi(2021)Wang and Qi]{StrongAugContrast21}
Wang, X. and Qi, G.
\newblock Contrastive learning with stronger augmentations.
\newblock \emph{CoRR}, abs/2104.07713, 2021.

\bibitem[Wang et~al.(2019)Wang, Sun, Liu, Sarma, Bronstein, and Solomon]{DGCNN}
Wang, Y., Sun, Y., Liu, Z., Sarma, S.~E., Bronstein, M.~M., and Solomon, J.~M.
\newblock Dynamic graph {CNN} for learning on point clouds.
\newblock \emph{ACM Trans. Graph.}, 38\penalty0 (5):\penalty0 146:1--146:12,
  2019.

\bibitem[Wang et~al.(2022)Wang, Yu, Rao, Zhou, and Lu]{P2P}
Wang, Z., Yu, X., Rao, Y., Zhou, J., and Lu, J.
\newblock {P2P:} tuning pre-trained image models for point cloud analysis with
  point-to-pixel prompting.
\newblock In \emph{Adv. Neural Inform. Process. Syst. (NeurIPS)}, 2022.

\bibitem[Wei et~al.(2022{\natexlab{a}})Wei, Fan, Xie, Wu, Yuille, and
  Feichtenhofer]{MaskFeat}
Wei, C., Fan, H., Xie, S., Wu, C.-Y., Yuille, A., and Feichtenhofer, C.
\newblock Masked feature prediction for self-supervised visual pre-training.
\newblock In \emph{IEEE/CVF Conf. Comput. Vis. Pattern Recog. (CVPR)},
  2022{\natexlab{a}}.

\bibitem[Wei et~al.(2022{\natexlab{b}})Wei, Wang, Schuurmans, Bosma, brian
  ichter, Xia, Chi, Le, and Zhou]{CoT22}
Wei, J., Wang, X., Schuurmans, D., Bosma, M., brian ichter, Xia, F., Chi,
  E.~H., Le, Q.~V., and Zhou, D.
\newblock Chain of thought prompting elicits reasoning in large language
  models.
\newblock In \emph{Adv. Neural Inform. Process. Syst. (NeurIPS)},
  2022{\natexlab{b}}.

\bibitem[Wu et~al.(2015)Wu, Song, Khosla, Yu, Zhang, Tang, and
  Xiao]{ModelNet15}
Wu, Z., Song, S., Khosla, A., Yu, F., Zhang, L., Tang, X., and Xiao, J.
\newblock 3d shapenets: A deep representation for volumetric shapes.
\newblock In \emph{IEEE/CVF Conf. Comput. Vis. Pattern Recog. (CVPR)}, pp.\
  1912--1920, 2015.

\bibitem[Wu et~al.(2018)Wu, Xiong, Yu, and Lin]{InstanceDiscrimination18}
Wu, Z., Xiong, Y., Yu, S.~X., and Lin, D.
\newblock Unsupervised feature learning via non-parametric instance
  discrimination.
\newblock In \emph{IEEE/CVF Conf. Comput. Vis. Pattern Recog. (CVPR)}, pp.\
  3733--3742. Computer Vision Foundation / {IEEE} Computer Society, 2018.

\bibitem[Xiang et~al.(2020)Xiang, Ding, and Han]{MultiExpertKD}
Xiang, L., Ding, G., and Han, J.
\newblock Learning from multiple experts: Self-paced knowledge distillation for
  long-tailed classification.
\newblock In \emph{Eur. Conf. Comput. Vis. (ECCV)}, volume 12350 of
  \emph{Lecture Notes in Computer Science}, pp.\  247--263. Springer, 2020.

\bibitem[Xie et~al.(2020)Xie, Gu, Guo, Qi, Guibas, and Litany]{PointContrast20}
Xie, S., Gu, J., Guo, D., Qi, C.~R., Guibas, L.~J., and Litany, O.
\newblock Pointcontrast: Unsupervised pre-training for 3d point cloud
  understanding.
\newblock In \emph{Eur. Conf. Comput. Vis. (ECCV)}, volume 12348 of
  \emph{Lecture Notes in Computer Science}, pp.\  574--591. Springer, 2020.

\bibitem[Xie et~al.(2022{\natexlab{a}})Xie, Geng, Hu, Zhang, Hu, and
  Cao]{DarkMIM22}
Xie, Z., Geng, Z., Hu, J., Zhang, Z., Hu, H., and Cao, Y.
\newblock Revealing the dark secrets of masked image modeling.
\newblock \emph{CoRR}, abs/2205.13543, 2022{\natexlab{a}}.

\bibitem[Xie et~al.(2022{\natexlab{b}})Xie, Zhang, Cao, Lin, Wei, Dai, and
  Hu]{ScaleMIM22}
Xie, Z., Zhang, Z., Cao, Y., Lin, Y., Wei, Y., Dai, Q., and Hu, H.
\newblock On data scaling in masked image modeling.
\newblock \emph{CoRR}, abs/2206.04664, 2022{\natexlab{b}}.

\bibitem[Yi et~al.(2016)Yi, Kim, Ceylan, Shen, Yan, Su, Lu, Huang, Sheffer, and
  Guibas]{ShapeNetPart16}
Yi, L., Kim, V.~G., Ceylan, D., Shen, I.-C., Yan, M., Su, H., Lu, C., Huang,
  Q., Sheffer, A., and Guibas, L.
\newblock A scalable active framework for region annotation in 3d shape
  collections.
\newblock \emph{ACM Trans. Graph.}, 35\penalty0 (6):\penalty0 1--12, 2016.

\bibitem[Yu et~al.(2022{\natexlab{a}})Yu, Wang, Vasudevan, Yeung,
  Seyedhosseini, and Wu]{CoCa22}
Yu, J., Wang, Z., Vasudevan, V., Yeung, L., Seyedhosseini, M., and Wu, Y.
\newblock Coca: Contrastive captioners are image-text foundation models.
\newblock \emph{Trans. Mach. Learn. Res. (TMLR)}, 2022{\natexlab{a}}.
\newblock ISSN 2835-8856.

\bibitem[Yu et~al.(2022{\natexlab{b}})Yu, Tang, Rao, Huang, Zhou, and
  Lu]{PointBERT}
Yu, X., Tang, L., Rao, Y., Huang, T., Zhou, J., and Lu, J.
\newblock Point-bert: Pre-training 3d point cloud transformers with masked
  point modeling.
\newblock In \emph{IEEE/CVF Conf. Comput. Vis. Pattern Recog. (CVPR)},
  2022{\natexlab{b}}.

\bibitem[Zbontar et~al.(2021)Zbontar, Jing, Misra, LeCun, and
  Deny]{BarlowTwins21}
Zbontar, J., Jing, L., Misra, I., LeCun, Y., and Deny, S.
\newblock Barlow twins: Self-supervised learning via redundancy reduction.
\newblock In \emph{Proc. Int. Conf. Mach. Learn. (ICML)}, volume 139 of
  \emph{Proceedings of Machine Learning Research}, pp.\  12310--12320. {PMLR},
  2021.

\bibitem[Zhang et~al.(2022{\natexlab{a}})Zhang, Chen, Zhang, Dong, and
  Ma]{CDS22}
Zhang, L., Chen, X., Zhang, J., Dong, R., and Ma, K.
\newblock Contrastive deep supervision.
\newblock In \emph{Eur. Conf. Comput. Vis. (ECCV)}, volume 13686 of
  \emph{Lecture Notes in Computer Science}, pp.\  1--19. Springer,
  2022{\natexlab{a}}.

\bibitem[Zhang et~al.(2022{\natexlab{b}})Zhang, Guo, Gao, Fang, Zhao, Wang,
  Qiao, and Li]{PointM2AE22}
Zhang, R., Guo, Z., Gao, P., Fang, R., Zhao, B., Wang, D., Qiao, Y., and Li, H.
\newblock Point-m2{AE}: Multi-scale masked autoencoders for hierarchical point
  cloud pre-training.
\newblock In \emph{Adv. Neural Inform. Process. Syst. (NeurIPS)},
  2022{\natexlab{b}}.

\bibitem[Zhang et~al.(2022{\natexlab{c}})Zhang, Guo, Zhang, Li, Miao, Cui,
  Qiao, Gao, and Li]{PointCLIP22}
Zhang, R., Guo, Z., Zhang, W., Li, K., Miao, X., Cui, B., Qiao, Y., Gao, P.,
  and Li, H.
\newblock Pointclip: Point cloud understanding by {CLIP}.
\newblock In \emph{IEEE/CVF Conf. Comput. Vis. Pattern Recog. (CVPR)},
  2022{\natexlab{c}}.

\bibitem[Zhang et~al.(2021)Zhang, Girdhar, Joulin, and Misra]{DepthContrast21}
Zhang, Z., Girdhar, R., Joulin, A., and Misra, I.
\newblock Self-supervised pretraining of 3d features on any point-cloud.
\newblock In \emph{Int. Conf. Comput. Vis. (ICCV)}, pp.\  10232--10243. {IEEE},
  2021.

\bibitem[Zhou et~al.(2022)Zhou, Wei, Wang, Shen, Xie, Yuille, and Kong]{iBoT}
Zhou, J., Wei, C., Wang, H., Shen, W., Xie, C., Yuille, A.~L., and Kong, T.
\newblock ibot: Image {BERT} pre-training with online tokenizer.
\newblock In \emph{Int. Conf. Learn. Represent. (ICLR)}, 2022.

\end{thebibliography}
}

\newpage
\appendix
\onecolumn
\begin{figure*}[ht]
    \begin{center}
    \includegraphics[width=0.9\linewidth]{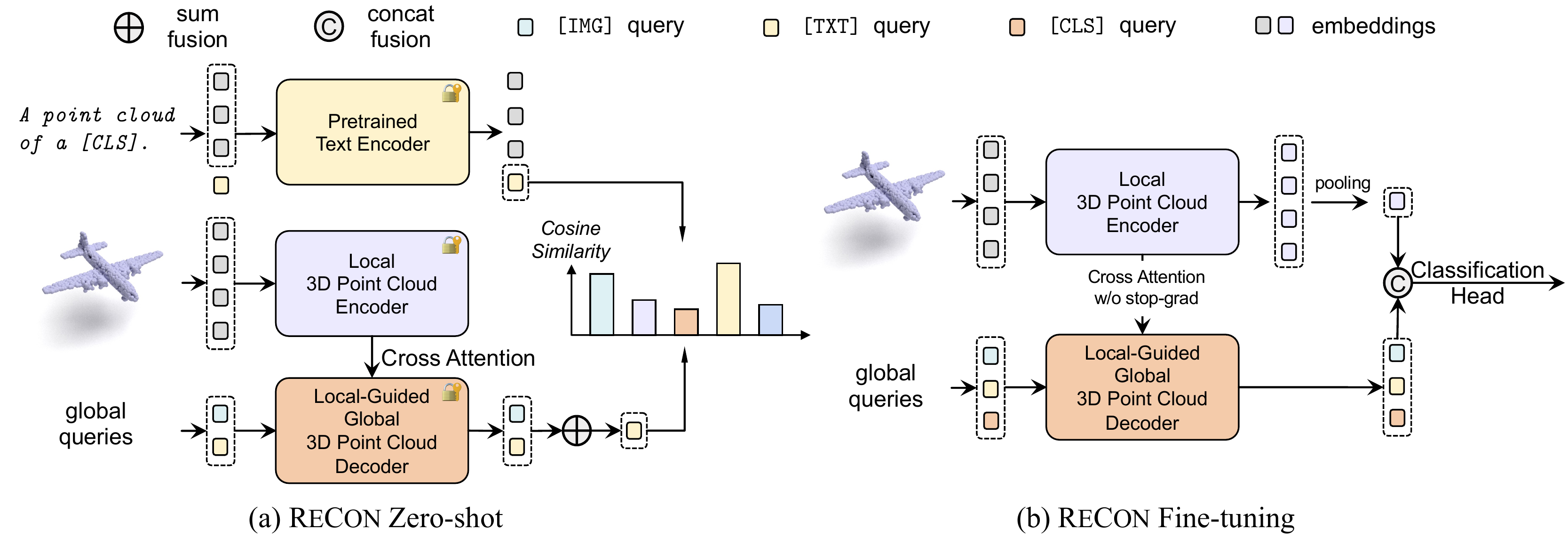}
    \caption{\textbf{Pipeline of \recon\ executing zero-shot and fine-tuning}. We fuse the pretrained global image (\texttt{[IMG]}) and text (\texttt{[TXT]}) query features by summation for zero-shot prediction. During fine-tuning, a new \texttt{[CLS]} token is added and fused with pretrained global \texttt{[IMG]} and \texttt{[TXT]} queries by concatenation for fine-tuning prediction. No stop-grad is used in CA connections during fine-tuning.}\label{fig:zeroshot}
    \end{center}
    \vspace{-10pt}
\end{figure*}
\section{Additional Related Works}
\textbf{Contrastive-Generative Representation Learning}~
In 2D vision and NLP, some works have been proposed for contrastive-generative representation learning. \citet{iBoT} propose to use an online tokenizer to distill local tokens and global class tokens, respectively. 
\citet{sim22} further proposes to perform three tasks at the same time: reconstruction, intra-view matching, and intra-image contrastive learning. 
This paradigm is also trending in the vision-language (VL) community, \citet{albef21} propose ALBEF that uses contrastive learning before modality fusion to make the reconstructed features encoded with more semantics. Through cross attention of different modalities, CoCa~\citep{CoCa22} fuses visual-language features while performing mask language modeling, where contrastive learning aligns the multi-modal global features. Recently, \citet{FLIP22} propose to use masked signals for contrastive VL learning, which greatly improves the training efficiency without losing performance. 
Different from these methods, our \recon\ is built for 3D representation learning, which is faced with a unique and serious \textit{low-data} challenge. 
Besides, we propose a novel \recon-block that models contrastive-generative representation learning as a generative pretraining \textit{guided} contrastive learning.
\begin{center}
\begin{table}[h!]
\vspace{-12pt}
\caption{\textbf{Training recipes for pretraining and downstream fine-tuning}.}
\label{tab:hyper_params}
\vskip 0.10in
\centering
\begin{tabular}{lcccc}
 & \texttt{Pretraining} & \multicolumn{2}{c}{\texttt{Classification}} & \texttt{Segmentation}\\
 \toprule[0.95pt]
 Config & ShapeNet & ScanObjectNN & ModelNet & ShapeNetPart\\
 \midrule[0.6pt]
 optimizer & AdamW & AdamW & AdamW & AdamW\\
 learning rate & 5e-4 & 2e-5 & 1e-5 & 2e-4 \\
 weight decay & 5e-2 & 5e-2 & 5e-2 & 5e-2 \\
 learning rate scheduler & cosine & cosine & cosine & cosine \\
 training epochs & 300 & 300 & 300 & 300\\
 warmup epochs & 10 & 10 & 10 & 10\\
 batch size & 128 & 32 & 32 & 16\\
 drop path rate & 0.1 & 0.2 & 0.2 & 0.1 \\
 \midrule[0.6pt]
 image resolution & 224$\times$224 & - & - & -\\
 image patch size & 16 & - & - & -\\
 number of points & 1024 & 2048 & 1024/8192 & 2048 \\
 number of point patches & 64 & 128 & 64/512 & 128 \\
 point patch size & 32 & 32 & 32 & 32 \\
 \midrule[0.6pt]
 augmentation & Rotation & Rotation & Scale\&Trans & - \\
 \midrule[0.6pt]
 GPU device & PH402 SKU 200 & RTX 2080Ti & RTX 2080Ti & RTX 2080Ti \\
\bottomrule[0.95pt]
\end{tabular}
\end{table}
\end{center}

\section{Additional Implementation Details}\label{app:impl_detail}
\subsection{Loss Function}
In this section, we detail the loss functions in our implementation of \textit{contrastive} models, \textit{generative} models, and our \recon\ models. Given a randomly sampled multimodal minibatch $\{\mathbf{p}_k, \mathbf{i}_k, \mathbf{t}_k\}_{k=1}^{N}$ with $N$ paired samples, where $\mathbf{p}_k$ is the $k$-th 3D point cloud sample, and $\mathbf{i}_k$, $\mathbf{t}_k$ are the corresponding multimodal views, \ie, rendered RGB image (rendered from the single view of the default pose) and text category description (\eg, \texttt{chair}), respectively.
By dividing the text category descriptions into $K$ groups $\{T_k\}_{k=1}^{K}$ where $T_k$ is the $k$-th unique text description, we obtain a fine-grained label set $\{\mathbf{y}_k = \ell : \mathbf{t}_k = T_\ell\}_{k=1}^{N}$ of ShapeNet. Hence, the minibatch becomes $\{\mathbf{p}_k, \mathbf{i}_k, \mathbf{t}_k, \mathbf{y}_k\}_{k=1}^{N}$, and we can use part of them or all of them to perform different self-supervised representation learning methods.

\textbf{Single-Modal Contrastive Loss}~
For the single-modal contrastive models (SMC), we use a \textit{supervised contrastive loss} following~\citet{SCL20}, where the supervision is generalized from the constructed label set.
Let $k\in \mathcal{K} \equiv \{1, \dots, N\}$ be the index of the minibatch samples where $\mathcal{A}_k = \mathcal{K}\backslash\{k\}$ is the index set for all samples other than the $k$-th \textit{anchor} sample. $\mathcal{P}_k \equiv \{p\in \mathcal{A}_k:\mathbf{y}_p=\mathbf{y}_k\}$ is the index for the \textit{positive} samples, and other samples with different labels are considered as the \textit{negative} samples.
Given the 3D student network $\mathcal{F}^{\text{P}}(\cdot)$, where $z^{\text{P}}_{\ell} = \mathcal{F}^{\text{P}}(\mathbf{p}_\ell), \ell \in \mathcal{K}$.
The loss $\mathcal{L}^{\text{SMC}}$ can be written as:
\begin{equation}\label{eq:smc_loss}
    \mathcal{L}^{\text{SMC}} = 
    \sum_{k\in\mathcal{K}} \frac{-1}{|\mathcal{P}_k|} \sum_{p\in\mathcal{P}_k}
    \log \frac{\exp(z^{\text{P}}_k \cdot z^{\text{P}}_p / \tau)}{\mathop{\sum}\limits_{a\in \mathcal{A}_i}\exp (z^{\text{P}}_k\cdot z^{\text{P}}_a / \tau)},
\end{equation}
where $\tau \in \mathbb R$ is the scalar temperature parameter which is set as 0.1~\citep{SCL20} for all contrastive models.

\textbf{Cross-Modal Contrastive Loss}~
For the cross-modal contrastive models (CMC), we use the contrastive InfoNCE loss~\citep{InfoNCE} following~\citet{CLIP}.
Given the minibatch, we can construct two groups of $N\times N$ pairs, \ie, point-image pairs $\big\{\{\mathbf{p}_m, \mathbf{i}_n\}: \forall m\in\mathcal{K}, \forall n\in\mathcal{K}\big\}$ and point-text pairs $\big\{\{\mathbf{p}_m, \mathbf{t}_n\}: \forall m\in\mathcal{K}, \forall n\in\mathcal{K}\big\}$.
Then the encoded representation of another modality from the same sample is used as \textit{positive}, while the others in the minibatch are used as \textit{negative} samples. 
Given a pretrained 2D teacher network $\mathcal{F}^{\text{I}}(\cdot)$ and a pretrained language teacher network $\mathcal{F}^{\text{T}}(\cdot)$, where $z^{\text{I}}_{\ell} = \mathcal{F}^{\text{I}}(\mathbf{i}_\ell), \ell \in \mathcal{K}$ and $z^{\text{T}}_{\ell} = \mathcal{F}^{\text{T}}(\mathbf{t}_\ell), \ell \in \mathcal{K}$ represent the decoded representations, respectively.
Similar to Eq. (\ref{eq:smc_loss}), the loss $\mathcal{L}^{\text{CMC}}$ is the summation of multimodal point-image loss $\mathcal{L}^{\text{CMC-PI}}$ and point-text loss $\mathcal{L}^{\text{CMC-PT}}$:
\begin{equation}\label{eq:cmc_loss}
    \mathcal{L}^{\text{CMC}} 
    =
    -\sum_{k\in\mathcal{K}}
    \left[
    \underbrace{\log \frac{\exp\left(z^{\text{P}}_k \cdot \texttt{stopgrad}(z^{\text{I}}_k) / \tau\right)}{\mathop{\sum}\limits_{a\in \mathcal{A}_i}\exp (z^{\text{P}}_k\cdot z^{\text{I}}_a / \tau)}}_{\mathcal{L}^{\text{CMC-PI}}} 
    +
    \underbrace{\log \frac{\exp\left(z^{\text{P}}_k \cdot \texttt{stopgrad}(z^{\text{T}}_k) / \tau\right)}{\mathop{\sum}\limits_{a\in \mathcal{A}_i}\exp (z^{\text{P}}_k\cdot z^{\text{T}}_a / \tau)}}_{\mathcal{L}^{\text{CMC-PT}}}
    \right],
\end{equation}
where $\tau \in \mathbb R$ is the scalar temperature parameter, and $\texttt{stopgrad}(\cdot)$ is the \textit{stop-gradient} operation. Here, no gradient is back-propagated to the image or text teachers, and hence the two cross-modal teachers are \textit{frozen}.

\textbf{Masked Point Modeling Reconstruction Loss}~
For the masked point modeling reconstruction (MPM), we use the $\ell_2$ Chamfer-Distance~\citep{ChamferDistance17} following~\citet{PointMAE}.
Denote $\mathcal{M}_\ell (\cdot),\ell\in\mathcal{K}$ as the masking operation. Let $\mathcal{R}_\ell\equiv \mathcal{F}^{\text{P}} \big(\mathcal{M}_\ell (\mathbf{p}_\ell)\big),\ell\in\mathcal{K}$ and $\mathcal{G}_\ell\equiv \mathbf{p}_\ell,\ell\in\mathcal{K}$ be the reconstructed point clouds and ground truth point clouds, respectively. 
The reconstruction loss $\mathcal{L}^{\text{MPM}}$ can be written as:
\begin{equation}\label{eq:mpm}
    \mathcal{L}^{\text{MPM}} =
    \sum_{k\in\mathcal{K}} 
    \left[
        \frac{1}{|\mathcal{R}_k|} \sum_{r\in\mathcal{R}_k}\mathop{\min}\limits_{g\in\mathcal{G}_k} \|r-g\|^2_2 +
        \sum_{g\in\mathcal{G}_k}\mathop{\min}\limits_{r\in\mathcal{R}_k} \|r-g\|^2_2
    \right].
\end{equation}

\textbf{\recon\ Loss}~
The \recon\ loss is the ensemble distillation as defined in \cref{eq:ensemble_kd}. The first term $\mathcal{L}^{\text{REC}}$ is the same as \cref{eq:mpm}, and the contrastive loss can be either single-modal contrastive loss defined in \cref{eq:smc_loss} or the cross-modal contrastive loss defined in \cref{eq:cmc_loss}.
Similar to SimSiam~\citep{SimSiam}, we find that the 3D student learning from the \textit{frozen} cross-modal teachers with \textit{stop-gradient} does not fall into the representation collapsing trap.
Therefore, we use the positive-only representation learning with Smooth $\ell_1$ loss $\text{Smooth-}\ell_1(\cdot,\cdot)$, which we find achieves the best performance (see Sec.~\ref{sec:ablation}). In this case, the cross-modal contrastive term $\mathcal{L}^{\text{CON}}$ can be written as follows:
\begin{equation}
    \mathcal{L}^{\text{CON}} = 
    \sum_{k\in\mathcal{K}}\Big[
    \text{Smooth-}\ell_1\big(z^{\text{P}}_k, \texttt{stopgrad}(z^{\text{I}}_k)\big) + \text{Smooth-}\ell_1\big(z^{\text{P}}_k, \texttt{stopgrad}(z^\text{T}_k)\big)\Big].
\end{equation}

\subsection{Experimental Details}
\textbf{Pretraining Details}~
We use ShapeNetCore from ShapeNet~\citep{ShapeNet15} as the pretraining dataset. ShapeNet is a clean set of 3D CAD object models with rich annotations, including $\sim$51K unique 3D models from 55 common object categories. 
For the generation of paired data, we take surface point samples of the 3D object model to generate 3D point clouds, the 3D models are lighted with textures for rendering 2D RGB images. 
Specifically, we use the MacOS Preview\footnote{\url{https://en.wikipedia.org/wiki/Preview_(macOS)}} to generate high-quality rendered images. 
The text description comprises category labels and manually designed prompt templates.
During pretraining, a unique image query \texttt{[IMG]} token and text query \texttt{[TXT]} token are trained to align the global representation from the image teacher and the text teacher, respectively. 
The overall pretraining includes 300 epochs, and we use a cosine learning rate~\citep{CosineLRSGDR} of 5e-4 warming up for 10 epochs. AdamW optimizer~\citep{AdamW19} is used, and the batch size is 128.
More details are shown in \cref{tab:hyper_params}.

\textbf{Downstream Transferring Details}~
\cref{fig:zeroshot} shows the pipeline when \recon\ transfers to downstream tasks, including zero-shot and fine-tuning. For zero-shot, we use simple summation to fuse multi-modal features, and the cosine similarity is used as the classification metric~\citep{CLIP}. 
During fine-tuning, we concatenate the pooled representation of local tokens and the learned global query tokens as model features.
For classification, we add a new global classification \texttt{[CLS]} token and concatenate it with the other queries before being fused to the classification head.
It is worth noting that due to the consistency of the optimization objective during fine-tuning, we cancel the stop gradient of the cross attention connections.
Without specifications, we report overall accuracy (OA) results without voting on the most challenging ScanObjectNN PB\_T50\_RS benchmark using 2,048 input points and ModelNet40 using 1,024 input points (1k P), and the zero-shot classification results are reported on the \textit{test} split.
More detailed training configurations are shown in \cref{tab:hyper_params}.

\textbf{Model Variants}~
\cref{tab:recon_variants} summarizes the \recon\ model configurations, which are grounded in a similar fashion of ViT variants~\citep{ViT}.
The default version ``\recon" (or \recon-Base) is directly adopted from previous works~\citep{PointMAE,ACT23}, except that the network is configured as two-stream rather than single-stream. We add the smaller ``Tiny" and ``Small" models, which have the same number of layers but with reduced channel dimension.
\vspace{-8pt}

\begin{table}[!t]
\caption{\textbf{Details of \recon\ model variants}. This table format follows~\citet{ViT}.} \label{tab:recon_variants}
\begin{center}
\begin{tabular}{lccccc}
\toprule[0.95pt]
Model & Layers & Hidden size & MLP size & Heads & \#Params\\
\midrule[0.6pt]
\br\recon-Tiny & 12 & 192 & 768 & 3 & 11.4M\\
\br\recon-Small & 12 & 256 & 1024 & 4 & 19.0M\\
\br\recon & 12 & 384 & 1536 & 6 & 43.6M \\
\bottomrule[0.95pt]
\end{tabular}
\end{center}
\vspace{-25pt}
\end{table}

\section{Additional Baselines}\label{app:add_baseline}
We show two additional simple fusion methods~\citep{MTLOverview}, including Vanilla Multi-task Learning and a Two-Tower network. Here, we make an analysis of these two baseline methods.

\begin{figure*}[t!]
    \begin{center}
    \includegraphics[width=0.9\linewidth]{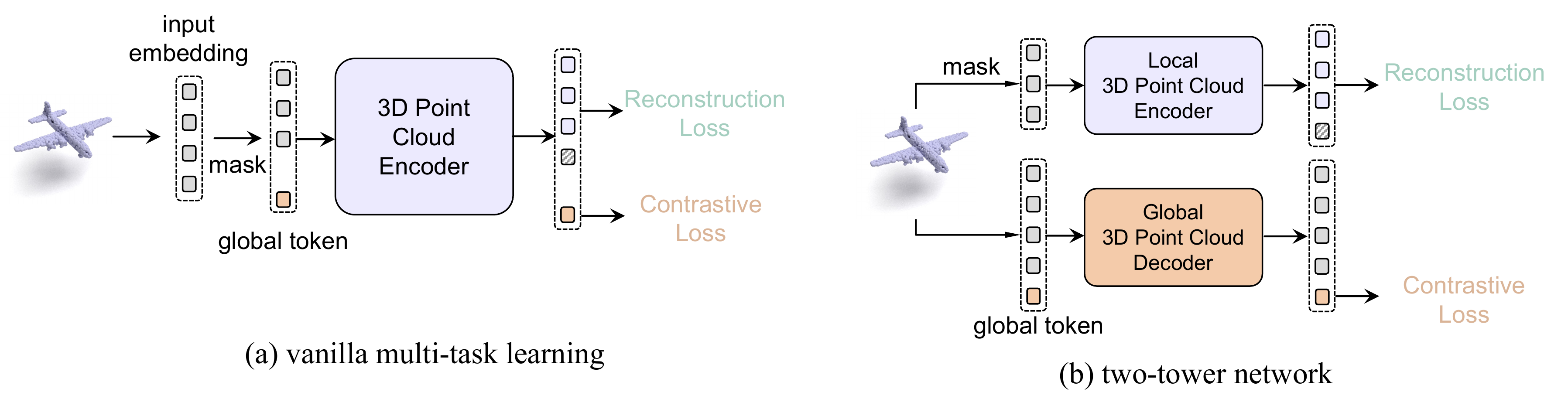}
    \vspace{-15pt}
    \caption{\textbf{Illustration of the vanilla multi-task learning and two-tower network baselines}.
    }\label{fig:add_baseline} 
    \end{center}
\end{figure*}

\textbf{Vanilla Multi-task Learning Fusion}~
As shown in \cref{fig:add_baseline}(a), Vanilla Multi-task Learning directly shares a standard Transformer as the encoder. The input embedding tokens take masked reconstruction as the pretext task, and the global tokens take the global contrast as the pretext task. Vanilla Multi-task Learning doesn't consider the pattern difference issue of the two tasks (see \cref{fig:data_scaling} and \cref{fig:atten_vis}). The transfer performances of Vanilla Multi-task Learning on ScanObjectNN and ModelNet40 are reported in \cref{tab:addition_baseline}. It is observed that this vanilla design leads to limited performance, which only improves the \textit{from scratch} OA by +0.89\% on ScanObjectNN, and no improvement on ModelNet40 is achieved.
This indicates task conflicts, and it is consistent with the analysis in Sec.~\ref{sec:intro} that it is \textit{non-trivial} for joint learning of these two tasks.

\textbf{Two-Tower Network}~
To verify whether the performance improvement of \recon\ comes from the form of a two-tower architecture, we design a simple Two-Tower network, shown in \cref{fig:add_baseline}(b). The Two-Tower network uses standard Transformers as the encoder for masked reconstruction and global contrastive learning, respectively.
During fine-tuning, it concatenates features from both streams for an ensemble (similar to \cref{fig:zeroshot}(b)). 
Clearly, the Two-Tower network doesn't suffer the \textit{pattern difference} issue.
The transfer performances of the Two-Tower network on ScanObjectNN and ModelNet40 are reported in \cref{tab:addition_baseline}. It can be seen that the Two-Tower network brings unsatisfactory performance, \ie, only +3.41\% and +0.5\% accuracy improvements by the \textit{from scratch} baseline. 
In comparison, our \recon\ uses the reconstruction task as guidance for global contrastive learning. 
As a result, \recon\ successfully disentangles the two tasks while preserving both merits, and significantly better improvements are achieved.
\begin{table}[!h]
\vspace{-15pt}
\caption{\textbf{Study of the additional baseline}. Overall accuracy (\%) without voting is reported.} \label{tab:addition_baseline}
\begin{center}
\begin{tabular}{lcc}
\toprule[0.95pt]
Method & ScanObjectNN & ModelNet40\\
\midrule[0.6pt]
Vanilla Multi-task Learning & 82.53 & 91.6\\
Two-Tower Network & 85.05 & 92.1\\
\rowcolor{linecolor}\recon & \textbf{90.63} & \textbf{94.1}\\
\bottomrule[0.95pt]
\end{tabular}
\end{center}
\vspace{-20pt}
\end{table}

\section{Additional Experiments}\label{app:add_exp}
\subsection{Additional Evaluations}\label{app:add_eva}
\vspace{-5pt}
\textbf{Linear SVM Evaluation}~
Linear SVM evaluation~\citep{MarginSVM92,SLTheory98} can be used to evaluate the discriminative quality of pretrained features~\citep{BenchmarkSSL19}. The results on ModelNet40 are shown in \cref{tab:linear}. It shows that our \recon\ outperforms Point-BERT, which also uses plain Transformers, by a clear margin of +6.0\%. 
Compared to methods using hierarchical Transformers, our \recon\ outperforms PointM2AE~\citep{PointM2AE22} by +0.5\%.
\begin{table}[h!]
\vspace{-10pt}
\caption{\textbf{Linear SVM classification on ModelNet40}. Overall accuracy (\%) without voting is reported.} \label{tab:linear}
\begin{center}
\begin{tabular}{lcc}
\toprule[0.95pt]
Method & Hierachical & ModelNet40\\
\midrule[0.6pt]
\bv Point-BERT~\citep{PointBERT} & $\times$ & 87.4\\
\bs OcCo~\citep{OcCo} & $\checkmark$ & 89.2\\
\bs CrossPoint~\citep{CrossPoint22} & $\checkmark$ & 91.2\\
\bh PointM2AE~\citep{PointM2AE22} & $\checkmark$ & 92.9\\
\rowcolor{linecolor}\br \recon  & $\times$ & \textbf{93.4}\\
\bottomrule[0.95pt]
\end{tabular}
\end{center}
\vspace{-10pt}
\end{table}

\textbf{3D Part Segmentation}~
To evaluate the geometric understanding performance within objects, we conduct the part segmentation experiment on ShapeNetPart~\citep{ShapeNetPart16}. 
Specifically, we concatenate the cross-modal feature of \recon\ into the global feature and use the same segmentation head as Point-MAE for a fair comparison. 
From \cref{tab:partseg}, it can be observed that \recon\ improves the \textit{from scratch} baseline by +1.4\% and +1.7\% Cls. mIoU and Inst. mIoU, respectively. 
Besides, \recon\ outperforms the SSL counterpart Point-MAE~\citep{PointMAE} by +0.4\% Inst. mIoU. 
It shows that the cross-modal global knowledge from \recon~cross-modal pretraining can still play a certain role in part segmentation.
\begin{center}
\begin{table}[h!]
\vspace{-5pt}
\caption{\textbf{Part segmentation on ShapeNetPart dataset}. The mIoU over all classes (Cls.) and the mIoU over all instances (Inst.) are reported.
$^\dagger$ denotes results with our proposed \br \recon-block built backbone architecture.
}
\label{tab:partseg}
\centering
\begin{tabular}{lcc}
\toprule[0.95pt]
Methods & Cls. mIoU (\%) & Inst. mIoU (\%)\\
\midrule[0.6pt]
\bs PointNet~\citep{PointNet}  & 80.4 & 83.7\\
\bs PointNet++~\citep{PointNet++} & 81.9 & 85.1\\
\bs DGCNN~\citep{DGCNN} & 82.3 & 85.2\\
\bs PointMLP~\citep{PointMLP} & 84.6 & 86.1\\
\midrule[0.6pt]
\bv Transformer~\citep{AttentionIsAllYouNeed} & 83.4 & 84.7\\
\br Transformer$^\dagger$~\citep{AttentionIsAllYouNeed} & 83.6 & 85.2\\
\bs PointContrast~\citep{PointContrast20} & - & 85.1\\
\bs CrossPoint~\citep{CrossPoint22} & - & 85.5\\
\bv Point-BERT~\citep{PointBERT} & 84.1 & 85.6\\
\bv Point-MAE~\citep{PointMAE} & - & 86.1\\
\br Point-MAE$^\dagger$~\citep{PointMAE} & 84.4 & 86.1\\
\bv ACT~\citep{ACT23} & 84.7 & 86.1\\
\rowcolor{linecolor}\br\recon & \textbf{84.8} & \textbf{86.4} \\
\bottomrule[0.95pt]
\end{tabular}
\end{table}
\end{center}

\textbf{Zero-Shot Recognition on Real-World Dataset}~
In \cref{tab:zeroshot_scan}, we show the zero-shot evaluation results on the real-world ScanObjectNN dataset. 
Though our \recon\ is pretrained on the synthetic dataset ShapeNet, it outperforms PointCLIP and CLIP2Point, which leverage depth images, by a clear margin. 
For example, on the most challenging PB\_T50\_RS benchmark, \recon\ achieves a Top-1 accuracy of 30.5\%, which is +7.2\% and +15.1\% higher than CLIP2Point and PointCLIP, respectively.  

\begin{table}[!h]
\caption{\textbf{Zero-shot 3D object classification on ScanObjectNN dataset}. Top-1 accuracy (\%) is reported. Ensemb. denotes whether to use the ensemble strategy with multiple text inputs.} \label{tab:zeroshot_scan}
\begin{center}
\begin{tabular}{lccccc}
\toprule[0.95pt]
Method  & Backbone & Ensemb. & OBJ\_ONLY & OBJ\_BG & PB\_T50\_RS\\
\midrule[0.6pt]
\bs PointCLIP~\cite{PointCLIP22} & ResNet-50 & $\times$ & 21.3 & 19.3 & 15.4 \\
\bv CLIP2Point~\cite{CLIP2Point22} & Transformer & $\checkmark$ & 35.5 & 30.5 & 23.3\\
\rowcolor{linecolor1}\br\recon & Transformer & $\times$ & \textbf{39.6} & \textbf{38.0} & \textbf{29.5}\\
\rowcolor{linecolor}\br\recon & Transformer & $\checkmark$ & \textbf{43.7} & \textbf{40.4} & \textbf{30.5} \\
\bottomrule[0.95pt]
\end{tabular}
\end{center}
\end{table}

\subsection{Additional Ablation Study}\label{app:add_ablation}
\textbf{Data Augmentation}~
We study the impact of using different data augmentations (DA) when fine-tuning \recon\ pretrained models on different downstream tasks, as shown in \cref{tab:augmentation}. The results show that \textit{Rotation} has the best performance improvement on ScanObjectNN, and \textit{Scale\&Translate} has the highest performance improvement on ModelNet40.
Therefore, we use by default \textit{Rotation} on ScanObjectNN and \textit{Scale\&Translate} on ModelNet40 if without specifications.
\begin{table}[!h]
\caption{
\textbf{Ablation study of data augmentations (DA) during fine-tuning}.
We report the fine-tuning overall accuracy (\%) without voting of \recon\ pretrained models.} \label{tab:augmentation}
\begin{center}
\begin{tabular}{lcc}
\toprule[0.95pt]
DA Strategy & ScanObjectNN & ModelNet40\\
\midrule[0.6pt]
- & 87.44 & 93.4\\
Scale \& Translate & 87.02 & \cellcolor{linecolor}\textbf{94.1}\\
Jitter & 90.22 & 92.9\\
Rotation & \cellcolor{linecolor}\textbf{90.63} & 92.3\\
Dropout & 87.40 & 93.5\\
Horizontal Flip & 87.27 & 93.6\\
\bottomrule[0.95pt]
\end{tabular}
\end{center}
\vspace{-10pt}
\end{table}

\textbf{Paired Data Ablation}~
During pretraining, we use the point cloud, rendered image, and text inputs generated from ShapeNet~\citep{ShapeNet15}, which makes the data contain clear matching attributes. 
To verify the dependence of \recon\ on paired data, we shuffle the rendered images under the same category for pretraining. 
The results are shown in \cref{tab:nonpair}. 
It shows that \recon\ has a performance degradation of less than 1\% in fine-tuning tasks and 6.4\% in the zero-shot tasks. 
\begin{table}[!h]
\caption{
\textbf{Ablation study on the paired data during pretraining}. Paired denotes the paired 3D point clouds, rendered images, and category text descriptions; unpaired data denotes data with shuffled images under the same category.
Overall accuracy (\%) without voting is reported. ModelNet40-FT represents the fine-tuning accuracy, and ModelNet40-ZS is the zero-shot result on the \textit{train}+\textit{test} split.} \label{tab:nonpair}
\begin{center}
\begin{tabular}{cccc}
\toprule[0.95pt]
Paired &  ScanObjectNN & ModelNet40-FT & ModelNet40-ZS\\
\midrule[0.6pt]
$\times$ & 90.22 & 93.8 & 60.4\\
$\checkmark$ & 90.63 & 94.5 & 66.8\\
\bottomrule[0.95pt]
\end{tabular}
\end{center}
\end{table}

\textbf{Prompt Ablation}~
For zero-shot evaluation, we construct the prompt by concatenating the prefix prompt \texttt{[P]} with the category text description \texttt{[C]}, followed by a suffix \texttt{[S]}, and the final prompt is \texttt{[P]}+\texttt{[C]}+\texttt{[C]}. 
For example, \texttt{A point cloud of a chair with white background}.
\cref{tab:prompts_ablation} shows the ablation study of results with only the prefix or suffix prompts we used, \ie, 20 prefixes and 4 suffixes.
Note that when we use the ensemble strategy, the results are ensembled by all the combinations of prefixes and suffixes.
The prompts can also be constructed from rendered images by powerful Vison-Language Foundation Models such as BLIP-2~\citep{BLIP223}, which may further bring improvements.
\begin{table}[!t]
\caption{
\textbf{Ablation study on the prompts for zero-shot learning}. \texttt{[C]} denotes the category text description, \texttt{[P]} denotes the prefix prompt, and \texttt{[S]} denotes the suffix prompt. Acc. (\%) represents the ModelNet40 zero-shot Top-1 Accuracy (\%).} \label{tab:prompts_ablation}
\begin{center}
\resizebox{0.9\linewidth}{!}{
\begin{tabular}{lclc}
\toprule[0.95pt]
\texttt{[P]}+\texttt{[C]} & Acc. (\%) & \texttt{[C]}+\texttt{[S]} & Acc. (\%)\\
\midrule[0.6pt]
\texttt{`'} + \texttt{[C]} & - & \texttt{[C]} +\texttt{`'} & -\\
\texttt{`A '} + \texttt{[C]} & 52.27 & \texttt{[C]} +\texttt{`.'} & 49.11\\
\texttt{`A model of '} + \texttt{[C]} & 53.04 & \texttt{[C]} +\texttt{` with white background.'} & 56.93\\
\texttt{`A model of a '} + \texttt{[C]} & 54.05 & \texttt{[C]} +\texttt{` with black context.'} & 60.57\\
\texttt{`An image of '} + \texttt{[C]} & 57.50 & - & -\\
\texttt{`An image of a '} + \texttt{[C]} & 56.36 & - & -\\
\texttt{`A 3D model of '} + \texttt{[C]} & 55.63 & - & -\\
\texttt{`A 3D model of a '} + \texttt{[C]} & 55.71 & - & -\\
\texttt{`A rendered model of '} + \texttt{[C]} & 56.48 & - & -\\
\texttt{`A rendered model of a '} + \texttt{[C]} & 56.52 & - & -\\
\texttt{`A point cloud of '} + \texttt{[C]} & 52.71 & - & -\\
\texttt{`A point cloud of a '} + \texttt{[C]} & 55.27 & - & -\\
\texttt{`A point cloud model of'} + \texttt{[C]} & 56.65 & - & -\\
\texttt{`A point cloud model of a '} + \texttt{[C]} & 54.46 & - & -\\
\texttt{`A 3D rendered model of '} + \texttt{[C]} & 55.83 & - & -\\
\texttt{`A 3D rendered model of a '} + \texttt{[C]} & 54.78 & - & -\\
\texttt{`A rendered image of '} + \texttt{[C]} & 58.79 & - & -\\
\texttt{`A rendered image of a '} + \texttt{[C]} & 56.48 & - & -\\
\texttt{`A 3D rendered image of '} + \texttt{[C]} & 59.07 & - & -\\
\texttt{`A 3D rendered image of a '} + \texttt{[C]} & 57.70 & - & -\\
\bottomrule[0.95pt]
\end{tabular}
}
\end{center}
\end{table}

\paragraph{Pretraining Learning Rate Ablation on Contrastive Models}
As discussed before, contrastive models can easily fall into the representation over-fitting trap. We find that it generally leads to a sensitivity to the pretraining learning rate. 
As shown in \cref{tab:lr_ablation}, when using the default learning rate for \recon\ pretraining (\ie, 5e-4), the contrastive model fails to generalize with unsatisfactory results. And when the learning rate is small, where the model learns slowly, a relatively better generalization performance is achieved with improved training stability (see \cref{fig:loss_curve} and the discussions).
We speculate that it is due to the \textit{low-data} challenge, which largely makes the model easily learn \textit{over-fitted} representations. As a result, the contrastive models are very sensitive to the pretraining learning rate, which has to be carefully adjusted. Hence, without specifications, we use this adjusted smaller learning rate for contrastive models.
\begin{table}[!h]
\caption{
\textbf{Ablation study on the pretraining learning rate (LR) of contrastive models}. We show the results of fine-tuned cross-modal contrastive models that are pretrained with different LRs. Overall accuracy (\%) without voting is reported.} \label{tab:lr_ablation}
\begin{center}
\begin{tabular}{cccc}
\toprule[0.95pt]
LR & ScanObjectNN & ModelNet40\\
\midrule[0.6pt]
5e-5 & 85.11 & 92.1\\
1e-4 & 86.49 & 92.4\\
5e-4 & 82.48& 90.3\\
1e-3 & 67.45 & 86.3\\
\bottomrule[0.95pt]
\end{tabular}
\end{center}
\end{table}

\paragraph{Ablation Study on Freezing Cross-Modal Teachers}
Unlike CLIP~\citep{CLIP}, \recon\ uses frozen cross-modal teachers in contrastive learning to acquire the dark knowledge of other modalities. We show the influence of freezing parameters on downstream tasks in \cref{tab:freeze_ablation}. It can be seen that if the model uses the InfoNCE~\citep{InfoNCE} loss function containing negative pairs, the impact of unfrozen parameters during pretraining is not significant. In contrast, for contrastive learning with only positive pairs, unfrozen parameters will cause serious performance degradation on ScanObjectNN and ModelNet.
\begin{table}[!h]
\caption{
\textbf{Ablation study on the freezing cross-modal teachers}. Freeze denotes whether both image and text teachers are frozen or not. Overall accuracy (\%) without voting is reported.} \label{tab:freeze_ablation}
\begin{center}
\begin{tabular}{cccc}
\toprule[0.95pt]
Freeze & Contrastive Metric & ScanObjectNN & ModelNet40\\
\midrule[0.6pt]
$\times$ & InfoNCE  & 89.87 & 93.7\\
$\times$ & Smooth $\ell_1$ & 84.77 & 91.2\\
\midrule[0.6pt]
$\checkmark$ & InfoNCE & 90.11 & 93.8\\
$\checkmark$ & Smooth $\ell_1$ & 90.63 & 94.1\\
\bottomrule[0.95pt]
\end{tabular}
\end{center}
\end{table}

\paragraph{Training Costs}
We report the singe-GPU without distributed training GPU hours and GPU memory usage of \recon\ and Point-MAE during pretraining and fine-tuning. We study \recon with three network configurations: \recon, \recon-Small, and \recon, which are described in \cref{tab:recon_variants}. Although the parameter count of \recon\ is almost doubled compared to Point-MAE, the memory consumption is mainly for storing intermediate variables of the model. However, \recon\ only adds three global queries to its intermediate variables compared to Point-MAE. Therefore, the memory consumption of \recon\ has mostly stayed the same compared to Point-MAE.
\begin{table}[!h]
\caption{
\textbf{Training costs comparison}.
We report the single-GPU training GPU hours and GPU memory costs during pretraining on ShapeNet and fine-tuning on ScanObjectNN.} \label{tab:traincost}
\begin{center}
\resizebox{\linewidth}{!}{
\begin{tabular}{lcccccc}
\toprule[0.95pt]
Method & \#Params & GFLOPS & Pretrain GPU Hours & Fine-tune GPU Hours & GPU Memory & ScanObjectNN\\
\midrule[0.6pt]
\bv Point-MAE & 22.1M & 4.8 & \textbf{32.7h} & 5.3h & 7766MB & 85.18\\
\rowcolor{linecolor1}\br\recon-Tiny & 11.4M & \textbf{2.4} & 40.9h & \textbf{3.8h} & \textbf{4578MB} & 89.10\\
\rowcolor{linecolor2}\br\recon-Small & 19.0M & 3.2 & 46.3h & 4.5h & 4710MB & 89.52\\
\rowcolor{linecolor}\br\recon & 43.6M & 5.3 & 54.5h & 6.1h & 7906MB & \textbf{90.63}\\
\bottomrule[0.95pt]
\end{tabular}
}
\end{center}
\end{table}

\paragraph{Edge Device Deployments}
We deploy the models using ONNX\footnote{\url{https://onnx.ai/}} and tested it on several common edge devices, including laptops, pads, smartphones, and single chip microcomputers (SCM), all of which are tested on CPU. \cref{tab:deploy} shows the number of frames per second (FPS) the model can make with inputs from the ModelNet40 dataset, which uses 1K point clouds per sample. The results demonstrate that our \recon\ network is easy to be deployed on various edge devices. For example, our small variant \recon-Tiny is even more efficient that surpasses the widely used PointNet++.
\begin{table}[!h]
\caption{
\textbf{Real-time deployments on edge devices}.
We report the number of frames per second (FPS).} \label{tab:deploy}
\begin{center}
\resizebox{\linewidth}{!}{
\begin{tabular}{lcccccc}
\toprule[0.95pt]
\multirow{3}{*}{Device} & Macbook Air & HUAWEI MatePad Pro & Honor X30 Android & Raspberry Pi 3B\\
& Laptop & Pad & Smartphone & Single Chip Microcomputer\\
& Apple M2 Silicon & Hisilicon Kirin 990 & Qualcomm Snapdragon 695 & Broadcom BCM2837\\
\midrule[0.6pt]
\bs PointNet++ & 83.6 & 25.6 & 16.7 & 2.5\\
\bv Point-MAE & 58.7 & 17.8 & 12.8 & 2.1\\
\rowcolor{linecolor1}\br\recon-Tiny & \textbf{103.5} & \textbf{29.8} & \textbf{20.2} & \textbf{3.7}\\
\rowcolor{linecolor2}\br\recon-Small & 78.1 & 24.5 & 14.6 & 2.7\\
\rowcolor{linecolor}\br\recon & 51.7 & 15.6 & 8.1 & 1.8\\
\bottomrule[0.95pt]
\end{tabular}
}
\end{center}
\end{table}

\newpage
\section{Discussions on Cross-Modal Teachers and Multimodal Training}
In \cref{tab:teacher_study}, we conduct an ablation study on \textit{pretrained teachers} and \textit{multimodal data} during pretraining. 
This analysis clearly demonstrates that 
(i) \recon+SMC that leverages single-modal contrastive learning still exhibits excellent performance on downstream tasks without including any other modality data or pretrained teachers. It achieves an overall accuracy of 89.73\% on ScanObjectNN, which is +1.31\% better than the generative-only baseline Point-MAE and +0.72\% better than the generative method ACT that leverages a pretrained 2D teacher. It demonstrates that our design of generative guidance for contrastive modeling is \textit{critical} and \textit{essential} for combining the merits of these two paradigms, which already yields superior results compared to other methods and has well tackled the raised issues. 
(ii) \recon+CMC (\textit{from scratch}) uses cross-modal contrastive learning on multimodal data while without any pretrained teachers, further bringing an improvement of +0.59\% to a remarkable 90.32\% overall accuracy on ScanObjectNN. It demonstrates that multimodal data is \textit{beneficial} since 3D data are seriously lacking.
(iii) \recon+CMC uses cross-modal contrastive learning with both multimodal data and pretrained teachers, further leading to an improvement of +0.31\% on ScanObjectNN. It demonstrates that pretrained teachers from other modality data and the usage of multimodal data in \recon\ (\textit{not} for other methods) can indeed further improve the performance for tackling the data dessert issue.
(iv) Vanilla Multi-Task Learning and Two-Tower Network baselines that simply transfer cross-modal knowledge from pretraining weights do not produce satisfactory results. We speculate that this is due to the pattern differences issue demonstrated in \cref{fig:data_scaling}, which is precisely the motivation behind our \recon-block. In contrast, our \recon+CMC outperforms these two baselines by a large margin. This shows that the benefits do not merely come from pretrained teachers but rather the fact that \textit{\recon\ design is an effective framework that guides contrastive learning with generative modeling}. It also shows that pretrained teachers are \textit{not} all you need, and the benefits of pretrained teachers or multimodal data can \textit{not} be obtained without our proposed \recon.
\begin{table}[!t]
\caption{
\textbf{Ablation study on pretrained teachers and multimodal data}. Pretrained Teacher denotes whether a pretrained teacher is used, and Multimodal Data denotes whether multimodal data is used during pretraining. SMC and CMC denote single-modal and cross-modal contrastive modeling methods, respectively. All results except the Vanilla Multi-Task Learning and Two-Tower Network baselines are conducted with our proposed \br\recon-block built backbone architecture during fine-tuning for a fair comparison. Overall accuracy (\%) without voting is reported.} \label{tab:teacher_study}
\begin{center}
\begin{tabular}{lcccc}
\toprule[0.95pt]
Method & Pretrained Teacher & Multimodal Data & ScanObjectNN & ModelNet40\\
\midrule[0.6pt]
\multicolumn{4}{l}{\textit{Contrastive} Methods}\\
\midrule[0.6pt]
\br SMC Only & $\times$ & $\times$ & 81.70 & 91.2\\
\br CMC Only & $\checkmark$ & $\checkmark$& 82.48 & 91.4\\
\midrule[0.6pt]
\multicolumn{4}{l}{\textit{Generative} Methods}\\
\midrule[0.6pt]
\br Point-MAE~\citep{PointMAE} & $\times$ & $\times$& 88.42 & 93.5\\
\br ACT~\citep{ACT23} & $\checkmark$ & $\times$ & 89.01 & 93.5\\
\midrule[0.6pt]
\multicolumn{4}{l}{\textit{Generative} + \textit{Contrastive} Methods}\\
\midrule[0.6pt]
Vanilla Multi-task Learning & $\checkmark$ & $\checkmark$ & 82.53 & 91.6\\
Two-Tower Network & $\checkmark$ & $\checkmark$ & 85.05 & 92.1\\
\rowcolor{linecolor1}\br \recon + SMC & $\times$ & $\times$ & 89.73 & 94.0\\
\rowcolor{linecolor2}\br \recon + CMC (\textit{from scratch}) & $\times$ & $\checkmark$ & 90.32 & 94.0\\
\rowcolor{linecolor}\br \recon + CMC & $\checkmark$ & $\checkmark$ & \textbf{90.63} & \textbf{94.1}\\
\bottomrule[0.95pt]
\end{tabular}
\end{center}
\end{table}

\section{Limitations and Future Works}
\recon\ is a general multimodal representation learning framework that leverages both merits of contrastive and generative modeling, which is demonstrated effective in 3D but is also general to any other modalities. However, there are some limitations of \recon, which may be two-fold. (i) The first limitation may come from the multimodal data and domain. This paper mainly explores \recon\ in 3D representation learning, and future explorations on multimodal problems like 2D Vision-Language may be intriguing. (ii) Another limitation may come from the architecture design, \ie, the \recon-block proposed in this work. It is our future exploration to extend \recon\ to be architecture-agnostic. 

\section*{Broader Impact}
The proposed \recon\ is a general framework that can be used for not only 3D representation learning but also all multimodal learning problems.
For example, by leveraging large-scale multimodal data like from the web~\citep{LAION5B2022}, one may obtain a foundational VL \recon\ that shares a similar property of CLIP~\citep{CLIP} since a multimodal alignment contrastive learning is used.
Besides, with the rapid development of Large Language Models (LLMs), \recon\ may also enable the potential for leveraging LLM like ChatGPT~\citep{ChatGPT} for LLM-assisted multimodal understanding.
Since \recon\ successfully unifies generative and contrastive modeling in a decent fashion, future applications may also involve AI-generated content (AIGC) but with cross-modal discriminative capability. For example, \recon\ can be trained for generative modeling that could be extended to generate contents based on input text instructions or other modalities.
We hope this work could motivate and facilitate future explorations on representation learning with multimodal or low-data inputs, which is critical for AI deployments in real-life.
However, all the potential impacts of the aforementioned applications should be taken into consideration while developing AI systems in human society.

\end{document}